\documentclass[11pt]{article}

\usepackage[final]{acl}

\usepackage{times}
\usepackage{latexsym}
\usepackage[dvipsnames]{xcolor}

\usepackage{lipsum}
\usepackage{arydshln}

\usepackage[utf8]{inputenc}
\usepackage[T1]{fontenc}
\usepackage{colortbl}
\usepackage[table]{xcolor}

\usepackage{microtype}

\usepackage{inconsolata}

\usepackage{graphicx}
\usepackage{enumitem}
\usepackage[normalem]{ulem} 
\usepackage{xcolor} 
\usepackage{booktabs}
\usepackage{amssymb}
\usepackage{url}
\usepackage[most]{tcolorbox}
\usepackage{fancyvrb}
\usepackage{pifont} 
\usepackage{soul} 

\usepackage{siunitx}  

\usepackage{booktabs}
\usepackage{tabularx}
\usepackage{array}
\usepackage{pifont}
\usepackage{hyperref}

\newcolumntype{Y}{>{\raggedright\arraybackslash}X}

\usepackage{appendix}
\usepackage{cuted}
\title{NADI 2026: \\The Second Multidialectal Arabic Speech Processing Shared Task }

\author{
   \textbf{Peter Sullivan}$^\alpha$ 
\quad \textbf{Bashar Talafha}$^\alpha$ 
\quad \textbf{Ahmed Ashraf}$^\beta$
\quad \textbf{Fethi Bougares}$^{\gamma,\delta}$ \\ 
 \textbf{Haroun Elleuch}$^{\gamma,\delta}$ 
\quad \textbf{Chiyu Zhang}$^\alpha$  
\quad \textbf{AbdelRahim Elmadany}$^\alpha$ 
\quad \textbf{Youssef Mohamed}$^\epsilon$ \\
\textbf{Salima Mdhaffar}$^\gamma$ 
\quad \textbf{Yannick Est{\`e}ve}$^\gamma$
\quad \textbf{Mohamed Elhoseiny}$^\epsilon$
\quad \textbf{Hamzah Luqman}$^\beta$ \\
\textbf{Nizar Habash}$^\zeta$
\quad \textbf{Muhammad Abdul-Mageed}$^{\alpha,\eta}$ \\
$^\alpha$The University of British Columbia \quad 
$^\eta$Canada Research Chair in NLP and ML \quad \\%
$^\beta$King Fahd University of Petroleum \& Minerals 
$^\gamma$Avignon Universit\'e \quad 
$^\delta$ELYADATA \quad \\
$^\epsilon$King Abdullah University of Science \& Technology 
$^\zeta$NYU Abu Dhabi \\
\texttt{
  \{prsull@student.,a.elmadany@,muhammad.mageed@\}ubc.ca} 
}

\begin{document}
\maketitle

\begin{strip}
  \centering
  \includegraphics[width=\textwidth]{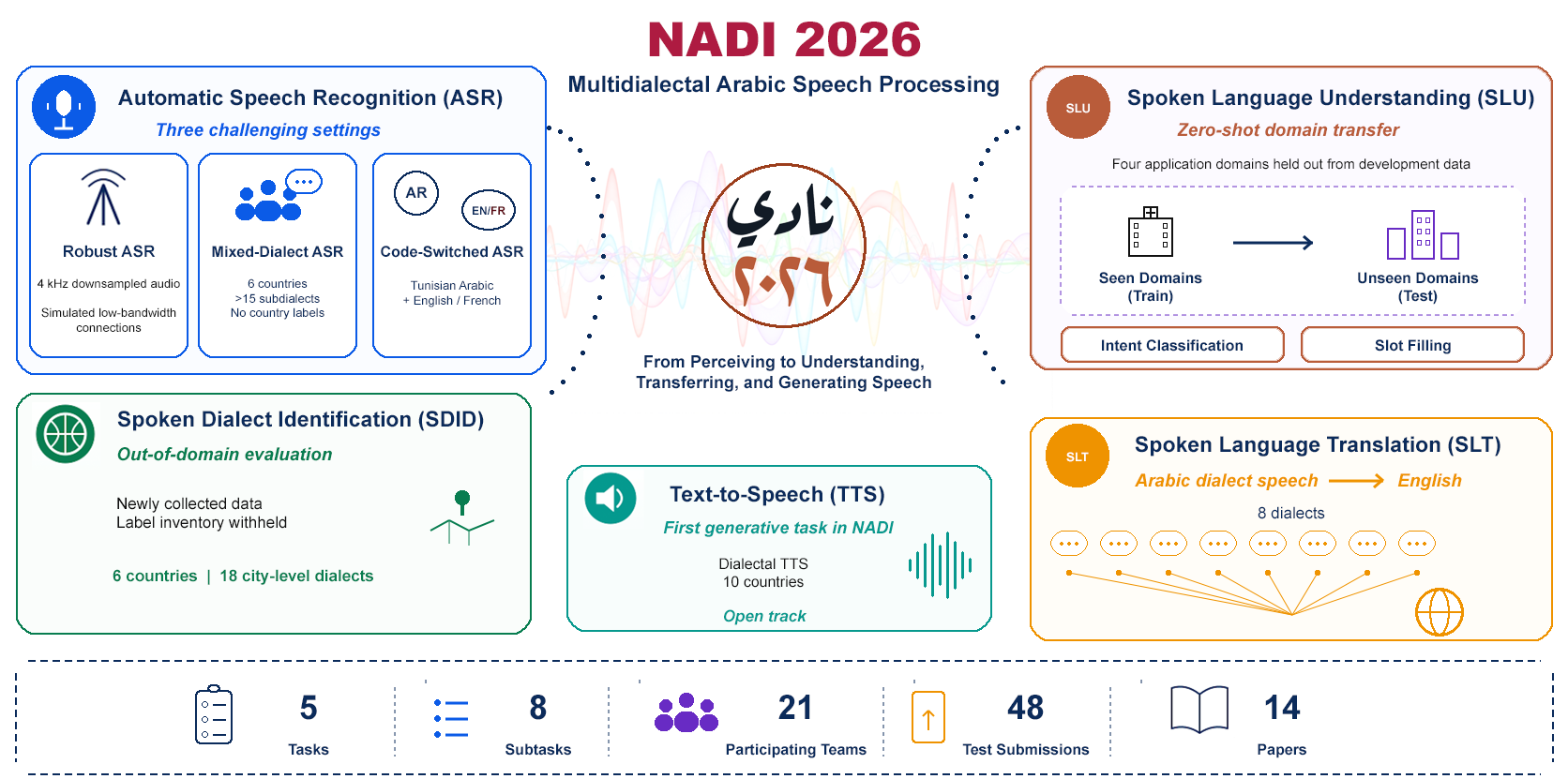}
  \vspace{-1.5em}
  \captionof{figure}{Overview of NADI 2026. The shared task spans the multidialectal Arabic spoken-language stack, covering speech perception through ASR and SDID, understanding through SLU, cross-language transfer through SLT, and speech generation through TTS. NADI 2026 further emphasizes realistic evaluation settings, including low-bandwidth, mixed-dialect, code-switched, out-of-domain, and zero-shot conditions. This edition comprises five tasks and eight subtasks, with 21 participating teams from at least 13 countries and 48 test-phase submissions.}
  \label{fig:participants_distri}
\end{strip}

\begin{abstract}
NADI 2026\footnote{\href{https://nadi.dlnlp.ai/2026/}{NADI 2026 Official Website: https://nadi.dlnlp.ai/2026/}} is the seventh edition of the Nuanced Arabic Dialect Identification (NADI) shared task series and the second dedicated to multidialectal Arabic speech processing. This edition comprises five tasks and eight subtasks spanning Automatic Speech Recognition (ASR), Spoken Dialect Identification (SDID), Text-to-Speech (TTS), Spoken Language Translation (SLT), and Spoken Language Understanding (SLU). NADI 2026 emphasizes realistic evaluation through low-bandwidth, mixed-dialect, code-switched, out-of-domain, and zero-shot settings, while introducing TTS, SLT, and SLU to the series for the first time. The shared task attracted $21$ participating teams from at least $13$ countries, with $48$ test-phase submissions, and $14$ submitted system-description papers. Results show that out-of-domain generalization remains a bottleneck and highlight the effectiveness of recent Arabic-specialized speech models, multimodal dialect identification, and ensemble methods. Overall, NADI 2026 provides a broader and more challenging benchmark for robust Arabic dialect speech processing.
\end{abstract}

\section{Introduction}

Modern speech systems report steadily improving performance on Arabic benchmarks~\cite{wang2024open},\footnote{Nearly a 10\% absolute WER drop from 2024's (nvidia/conformer-ctc-large-ar) and the newly released audarai/Audar-ASR-V1-Turbo.} yet speakers of Arabic dialects still meet systems that misidentify them, transcribe them poorly, or cannot speak to them in their own variety. Part of this gap is due to coverage: the International Organization for Standardization lists 28 active varieties under the ISO 639-3 Arabic macrolanguage code \texttt{ara},\footnote{\href{https://iso639-3.sil.org/code/ara}{https://iso639-3.sil.org/code/ara}} and even that figure understates city-to-city variation and the urban--rural divides. With few exceptions, benchmarks are usually assembled from clean, in-domain, and limited-variety recordings, whereas deployed systems encounter degraded audio, speakers whose variety is unknown in advance, utterances switching language or variety mid-sentence, and domains not anticipated at training time~\cite{sullivan-etal-2026-arab}. A system can look stronger under the first set of conditions (characteristic of most benchmarks) than on the second.  








Dialectal Arabic (DA) makes these challenges real. Unlike Modern Standard Arabic (MSA), the formal register for broadcast news and official addresses, DA is what speakers use in daily life and has no standard orthography~\cite{ali2019mgb,ali2015multi}. The same word is legitimately written in more than one way, making transcriptions inconsistent and \textit{evaluation} itself less reliable. DA is also morphologically complex~\cite{dahou2025survey}, and code-switching into English, French, or MSA is far from exceptional~\cite{dahou2025survey, besdouri2024arabic, ali2021connecting}. Compounding this, some of the largest transcribed Arabic speech corpora, such as MGB2~\cite{ali2016mgb} and QASR~\cite{mubarak-etal-2021-qasr}, are predominantly MSA. Dedicated dialectal corpora such as MASC~\cite{al-fetyani2022masc}, ADI-20~\cite{elleuch25_interspeech}, and Casablanca~\cite{talafha2024casablanca} have begun to close this gap. Although the arrival of these dialectal corpora is precisely what makes evaluations reported here possible, they remain small compared to what MSA and high-resource languages enjoy. 

The Nuanced Arabic Dialect Identification (NADI) shared task series was created to concentrate community efforts on this gap. It began in 2020 with country- and province level dialect identification over social media text~\cite{abdul2020nadi}, and across subsequent editions added variety identification across MSA and DA~\cite{abdul2021nadi}, dialectal sentiment analysis~\cite{abdul2022nadi}, dialect-to-MSA machine translation~\cite{abdul2023nadi}, and multi-label dialect identification with level-of-dialectness estimation~\cite{abdul2024nadi}. In 2025, NADI pivoted to speech, with a focus on spoken dialect identification, multidialectal automatic speech recognition (ASR), and spoken diacritic restoration~\cite{talafha-etal-2025-nadi}. Individual subtasks have come and gone, but dialect identification has run throughout the series: it is the tool that allows researchers to aggregate varieties, route utterances to dialect-specific models, and audit how well open resources cover the Arab world. Table~\ref{tab:nadi_series} summarizes this trajectory and we provide a detailed comparison in Appendix~\ref{tab:nadi_evolution_full}.

\begin{table*}[t]
\centering

\resizebox{0.85\textwidth}{!}{%
\begin{tabular}{
@{}
l
c c c c c
>{\columncolor{gray!10}}c
>{\columncolor{gray!20}[\tabcolsep][0pt]}c
@{}
}

\toprule
\textbf{Comparison dimension}
& \textbf{2020}
& \textbf{2021}
& \textbf{2022}
& \textbf{2023}
& \textbf{2024}
& \textbf{2025}
& \textbf{2026} \\

\midrule

Modality
& Text
& Text
& Text
& Text
& Text
& Speech
& Speech \\

\# Subtasks
& 2
& 4
& 2
& 3
& 3
& 3
& 8 \\

Task families
& T1
& T1
& T1, T2
& T1, T3
& T1, T3, T4
& T1, T5, T6
& T1, T5, T7--T9 \\

\midrule

Registered teams
& 61
& 53
& 41
& 58
& 51
& 44
& 65 \\

Participating teams
& 18
& 8
& 21
& 18
& 12
& 8
& 21 \\

Test submissions
& 56
& 68
& 105
& 76
& 76
& 100
& 48 \\

Accepted papers
& 14
& 7
& 15
& 13
& 8
& 7
& 14 \\


\bottomrule
\end{tabular}%
}

\caption{
Evolution of the NADI shared task series from 2020 to 2026.
The series focused on text-based tasks from 2020 to 2024, transitioned
to speech in 2025, and further expanded its speech-processing scope in 2026.
Task-family codes are normalized across editions.
\textbf{T1}: Dialect Identification;
\textbf{T2}: Sentiment Analysis;
\textbf{T3}: Dialect-to-MSA Machine Translation;
\textbf{T4}: Arabic Level of Dialectness Estimation;
\textbf{T5}: Automatic Speech Recognition;
\textbf{T6}: Diacritic Restoration;
\textbf{T7}: Text-to-Speech;
\textbf{T8}: Spoken Language Translation; and
\textbf{T9}: Spoken Language Understanding.
For a detailed comparison of task coverage, task characteristics,
geographic and variety scope, and evaluation metrics, see
Appendix Table~\ref{tab:nadi_evolution_full}.
}
\label{tab:nadi_series}

\end{table*}


NADI 2026 is the seventh NADI shared task and the second dedicated to speech. Its design follows from the argument above and moves along two axes. The first is \textit{realism}: rather than introducing new tasks, we re-introduce established ones under conditions that would break a system during deployment. ASR is evaluated on audio downsampled to 4~kHz and resampled to mimic low-bandwidth connections~\cite{bouchakour2025enhancing}, on mixed dialect speech~\cite{besdouri2024arabic} for which no country labels are supplied, and on Tunisian speech with frequent within-utterance switching into French or English. Spoken dialect identification (SDID) is evaluated out of domain~\cite{sullivan23_interspeech,abdullah25_interspeech}, on newly collected data whose label inventory is withheld from participants, closing the loophole that lets system exploit a known label set~\cite{sullivan23_interspeech,abdullah25_interspeech}.\footnote{The inventory comprises six countries and 18 city-level dialects.} Spoken language understanding reserves four application domains that appear nowhere the development data, thereby testing zero-shot domain transfer.  The second axis is \textit{scope}: with the addition of text-to-speech (TTS), spoken language translation (SLT), and spoken language understanding (SLU), NADI 2026 spans the full spoken-language stack for Arabic dialects. That is, perceiving speech (ASR, SDID), understanding it (SLU), transferring it across languages (SLT), and generating it (TTS). It is noteworthy that NADI 2026 introduces the first generative subtask in the history of the series, i.e., TTS.

NADI 2026 comprises five tasks and eight subtasks, all except TTS running on  Codabench. Across a development phase running from 16 June  to 20 July and a test phase from 20 to 28 July, we received 22 development-phase and 48 test-phase submissions, with 21 teams from at least 13 countries submitting to at least one subtask. We received and accepted 14 system description papers.

A number of findings across the various shared task leaderboards are worth mentioning. First, out-of-domain generalization is a challenging problem: our dialect identification baseline falls from 81.19\% accuracy in domain to 28.70\% out of domain, and the best participating system recovers only to 56.15\%. Second, no single recipe wins. For example, on code-switched ASR, a fully finetuned 2B parameter Arabic encoder--decoder, a LoRA adaptation training 3.6\% of the parameters of its nearest rival, and a ten-system ROVER ensemble finish within 0.12 WER of one another. Third, Arabic-centric foundation models were utilized to great success this year, being used in five of the top six SDID entries, the top entry of one of the ASR subtasks, and second place entry in the other two ASR subtasks. This last finding is a visible shift from NADI 2025, which was dominated by general-purpose multilingual models. Fourth, system combinations helped almost everywhere.

\noindent We offer the following contributions:

\begin{itemize}
    \item A \textit{realism-first benchmark} covering recognition, identification, understanding, translation, and synthesis, each posed under a  distribution shift rather than in-domain conditions.

    \item \textit{New evaluation sets spanning five tasks}: previously unreleased portions of Casablanca~\cite{talafha2024casablanca} including a low-bandwidth condition; a six-country, 18 city dialect identification set with withheld label inventory; dialectal TTS data covering ten countries; WhiteHouse~\cite{bougares2026whitehouse} evaluation data for dialect-to-English speech translation; and a new SLU evaluation set with four zero-shot domains. 

    \item \textit{A systematic analysis of the 21 participating systems}, 14 of which are documented in system description papers, identifying which adaptation, ensembling, and post-processing strategies generalize across subtasks and which do not.

\end{itemize}

The remainder of the paper is structured as follows: we provide a review of  literature relevant to the shared task in \S\ref{sec:lit}; give an overview of this year's subtasks in \S\ref{sec:nadi2026}; describe the baseline systems and team participation in \S\ref{sec:teams_baselines}; provide the results of each subtask in \S\ref{sec:results}; discuss key findings in \S\ref{sec:disc}; and finally we conclude in \S\ref{sec:conclusion}. 
\section{Literature Review}\label{sec:lit}
Low-resource ASR increasingly uses transfer learning~\cite{kopikar2026amchi}, self-supervised pretraining~\cite{zhuo2025vietasr}, and hybrid HMM/self-supervised architectures~\cite{klejch2025practitioner}. In-context learning offers another route for recognizing previously unseen languages~\cite{li2025context}, while large multilingual encoders have expanded coverage~\cite{alabi2024afrihubert}. These methods are particularly relevant to Arabic dialects, which lack the transcribed resources available for MSA~\cite{sullivan-etal-2026-arab,ali2017speech,madar-2018-bouamor}. 

The NADI shared tasks have consequently promoted research on dialectal and code-switched Arabic~\cite{talafha-etal-2025-nadi}. Because standard-language ASR systems often perform poorly on dialectal speech, recent Arabic research has investigated dialect-specific fine-tuning~\cite{wang2024open,ozyilmaz2025overcoming} and joint dialect identification and speech recognition~\cite{kumar2026jointly}. Spoken language identification has similarly progressed from i-vector~\cite{dehak2011language} and x-vector~\cite{snyder2018spoken} systems to Transformer-based models~\cite{vaswani2017attention,babu22_interspeech}. Arabic dialect identification followed this transition~\cite{ali2017speech,shon2017qcri,ali2019mgb,elleuch2025elyadata,talafha-etal-2025-nadi}, although weak cross-domain generalization remains a persistent limitation~\cite{sullivan23_interspeech,abdullah25_interspeech}.

Although DA speech processing has made strides in other areas, resource constraints pose a challenge for TTS~\cite{laouirine2024tunartts} and SLT~\cite{bougares2026whitehouse,agostinelli-etal-2025-findings,bougares2025tedxtn,hamed2022arzen} while SLU has mainly been focused on Tunisian Arabic~\cite{mdhaffar2024taric,mdhaffar2025sense}. 
We provide a more detailed review of related literature in Appendix ~\ref{sec:appendix_lit}.

\section{NADI 2026}\label{sec:nadi2026}

This year we include five tasks, with additional subtasks for ASR and SLU. Except for TTS, all subtasks were evaluated on the Codabench platform, with a development phase and an evaluation phase. The remainder of this section gives a breakdown of each of the tasks/subtasks.

\subsection{Task 1: ASR}
\paragraph{Task Description.} 

Our three ASR subtasks involve transcribing audio utterances from different dialectal, noisy, and domain conditions. The first of these subtasks, \textbf{1.1 Robust ASR}, emphasizes performance under a simulated low bandwidth scenario. The second, \textbf{1.2 Mixed Dialect ASR}, highlights the challenge of processing DA utterances without dialect labels. The final subtask, \textbf{1.3 Code-switched ASR} focuses on code-switching between Tunisian Arabic, English, and French. 

\paragraph{Data.}

For the Robust ASR subtask, we apply a downsampling strategy to audio samples from the Casablanca~\cite{talafha2024casablanca} dataset, with the test data coming from unreleased portions of the dataset. We downsample these files to 4khz and then resample them up to 16khz, in an effort to mimic extremely low bandwidth connections.

In the Mixed-Dialect subtask, we use newly collected data across six countries covering over 15 subdialects. No country labels are provided as part of the task.

Our Code-switched subtask targets transcribing spontaneous Tunisian Arabic speech with heavy French and English code-switching. Participants to this subtask are evaluated using newly collected data of around 1 hour and 50 minutes that represent 2,123 segments from a mix of talks and podcasts across diverse domains. 

\paragraph{Evaluation Metrics.}

We use word error rate (WER) as our primary metric and character error rate (CER) as a secondary metric. For the official rankings we calculate these error rates without hamza normalization. 

\subsection{Task 2: Spoken Dialect Identification}

\paragraph{Task Description.} 
SDID involves estimating which of a closed set of dialects a spoken utterance belongs to. These tools have been observed to overfit to training domains~\cite{sullivan23_interspeech,abdullah25_interspeech}, and thus this year we focus on out-of-domain SDID.

\paragraph{Data.}
Our target test data comes from newly collected data covering six countries and 18 city-level dialects within these six countries. Participants were not informed of the possible countries that would be included in the test data, except that they would be a subset of the 20 country labels used in ADI-20~\cite{elleuch25_interspeech}.

\paragraph{Evaluation Metrics.} We use two metrics for this task: Accuracy and the Average Cost metric from NIST Language Recognition Evaluation 2022~\cite{lee2022nist}.  

\subsection{Task 3: Dialectal Text-to-Speech}

\paragraph{Task Description.} 
This task requires participants to synthesize dialectal Arabic speech from text prompts, capturing the characteristic pronunciation, rhythm, and prosody of the target dialect. Newly introduced to NADI this year, the task aims to establish benchmarking practices for an area where standard evaluation remains scarce. Given only text prompts, systems are evaluated on generating high-quality speech across ten Arab countries: Egypt, Sudan, Saudi Arabia, the United Arab Emirates, Jordan, Palestine, Tunisia, Algeria, Morocco, and Yemen. To encourage realistic acoustic quality and prosodic fidelity across diverse dialects, this task was conducted as an open track, allowing participants to utilize any external training data.

\paragraph{Data.}
To build the evaluation dataset, we sampled sentences representing each target dialect from the Bulbul dataset~\citep{cite_bulbul}.The text in Bulbul was manually collected and verified by native speakers from each respective country. Details of the data processing can be found in Appendix ~\ref{sec:app_tts_data}.

\paragraph{Evaluation Metrics.}
To evaluate our task, we adopt two widely used neural non-intrusive metrics: UTMOSv2~\citep{baba2024utmosv2} and NISQA~\citep{Mittag_2021, mittag20_interspeech}, both of which leverage deep learning models trained on human perception ratings. To capture semantic performance, we also calculate WER and CER using the 7B parameter Omnilingual LLM-based ASR system with language set to `arb\_Arab'~\cite{omnilingual2025omnilingual}. 


\subsection{Task 4: Spoken Language Translation} 

\paragraph{Task Description.} 
Participants in this task are expected to train systems that translate from eight Arabic dialect speech inputs into English. Systems must cope with dialectal variation, spontaneous speech, and real-world speaking conditions.
\paragraph{Data.} Evaluation data is sourced from WhiteHouse ~\cite{bougares2026whitehouse}, the multi-dialectal Arabic speech translation dataset. WhiteHouse was created by supplementing the multi-dialectal Casablanca corpus with English translation. In total, the evaluation dataset comprises 6,788 and 6,780 utterances representing the validation and test sets of the eight Arabic dialects considered in the Casablanca dataset. Participants were provided with the WhiteHouse validation set for model development, while the test set, presented in Table \ref{tab:whitehouse_test} was used for the official evaluation.

\paragraph{Evaluation Metrics.}
All participant submissions are evaluated with BLEU~\cite{papineni2002bleu} and chrF~\cite{popovic-2015-chrf} scores. All submissions are evaluated with case-sensitivity and without punctuation. Scores are computed using the Sacrebleu toolkit~\cite{sacrebleu}.

\subsection{Task 5: Spoken Language Understanding} 
\paragraph{Task Description.}
SLU refers
to natural language processing tasks that aim
to extract semantic information from speech~\cite{tur2011spoken}.
The goal of this task is to extract semantic information directly from Tunisian dialect speech, beyond literal transcription under spontaneous, possibly code-switched conditions. We break this into two subtasks \textbf{5.1 Intent Recognition} and \textbf{5.2 Slot Filling}. 
For intent classification, participants must classify spoken utterances into predefined intent categories capturing the speaker's goal. In the slot filling subtask, participants must identify and extract semantic information (entities, attributes, task-relevant details) from spoken utterances.

\paragraph{Data.} 
Participants were provided with the original SLURP-TN~\cite{slurptn} training and
development sets for system development. 
For evaluation, we introduce a new dataset recorded following the same guidelines and annotation protocol as SLURP-TN, ensuring consistency in terms of task design, intent and slot definitions, and data collection procedures.
The new evaluation set contains 989 distinct utterances and amounts to one hour of clean
speech. See Appendix~\ref{sec:app_slu} for complete details.

\paragraph{Evaluation Metrics.} The SLU slot filling task was evaluated
based on terms of Concept Error Rate (CoER) and
Concept/Value Error Rate (CVER).
 CoER is computed similarly to WER by taking into
account only the semantic labels in the reference and hypothesis annotations. The CVER
computation is identical, but the occurrences
of concept/value pairs are taken into account
instead of the concept alone. CVER implies
that if any character within the word’s support
prediction or the concept tag prediction differs
from the reference, the entire prediction for
that concept is considered an error. 
CoER and CVER are detailed in~\cite{laperriere2022spoken}.

\section{Shared Task Teams \& Baselines} 
\label{sec:teams_baselines}

\subsection{Participating Teams}
 We provide an overview of participation across the phases in Table ~\ref{tab:participation}, with a final total of 21 participating teams listed in Table ~\ref{tab:teams}. 

\begin{table*}[!ht]
\centering
\small
\tabcolsep8pt
\begin{tabular}{ll c}
\toprule
\textbf{Team Name} & \textbf{Affiliation} & \textbf{Subtask}  \\
\midrule
\textbf{A3da2 Messi} & Global Brands Group (GBG) & 5.1 \\ 
\textbf{Abjad AI}~\cite{abjad-nadi2026} & Abjad AI, Jordan \& Saudi Arabia &  1.3, 2 \\
\textbf{AI Elites}~\cite{ai_elites-nadi2026} & KAUST Academy, Saudi Arabia  & 2 \\
\textbf{Ar\_India} & Independent Researchers, India & 1.1, 1.3, 2 \\
\textbf{ArabSpeech Minds} & Nile University, Egypt & 2\\
\textbf{Aslema}~\cite{aslema-nadi2026} & QCRI, Hamad Bin Khalifa University, Qatar  & 5.1, 5.2 \\
\textbf{CIS-OpenSLU}~\cite{cis_openslu-nadi2026} & Nile University, Egypt  &  5.1 \\
\textbf{CodeZone
}~\cite{codezone-nadi2026} & Lebanese university, Lebanon \\
& \verb|   |\& Vivekananda Global University, India &  2 \\
\textbf{FARABI-Fraunhofer-IAIS} & Fraunhofer IAIS, Germany & 2, 5.1 \\
\textbf{Itgan}~\cite{itgan-nadi2026} & Independent Researcher, UK & 1.1, 1.2, 1.3 \\
\textbf{Lynx}~\cite{lynx-nadi2026} & Mohammed VI Polytechnic University, Morocco & 2, 5.1 \\
\textbf{NAMAA
}~\cite{namaa-nadi2026} & NAMAA Community, Saudi Arabia &  1.2, 1.3, 2\\
\textbf{Nile}~\cite{nile-nadi2026} & Nile University, Egypt  & 1.1 \\
\textbf{KAND CA}~\cite{kand-nadi2026} & KAND CA, Canada  & 1.1, 1.2, 1.3, 2, 3, 4\\
\textbf{Resonate}~\cite{resonate-nadi2026} & Nile University, Egypt & 1.1, 2 \\
\textbf{Sabaa}~\cite{sabaa-nadi2026} & Najran University, Saudi Arabia & 1.1 \\
\textbf{Salesteq} & Salesteq, Switzerland &  1.1, 1.2, 2 \\
\textbf{Sense} & Birzeit University, Palestine & 4 \\
\textbf{Thakaa}~\cite{thakaa-nadi2026} & Thakaa, Advanced Company for Information\\
&  \verb|   |Technology, Saudi Arabia & 1.1, 1.3, 2 \\
\textbf{Wifaq}~\cite{wifaq-nadi2026} & Independent Researcher, Saudi Arabia  & 1.1 \\
\textbf{Zila} & Hacettepe University, Türkiye &  5.2 \\
\bottomrule
\end{tabular}
\caption{List of teams that registered and participated in NADI 2026 shared task. Teams with accepted papers are cited.}
\label{tab:teams}
\end{table*}

\subsection{Baselines}

We use the \texttt{whisper-large-v3} model~\cite{radford2023robust} as an off-the-shelf baseline for ASR and SLT tasks. For SDID we finetune an ECAPA-TDNN~\cite{desplanques20_interspeech} model on a subset of the ADI20 dataset~\cite{elleuch25_interspeech}. For the TTS task, we have generated speech samples using XTTS-v2 \footnote{\href{https://github.com/idiap/coqui-ai-TTS}{https://github.com/idiap/coqui-ai-TTS}} for all dialects. We choose only one reference speaker and there were no special prompting used.
For SLU, we provide Whisper-based systems for both subtasks. The Intent Recognition baseline is a \texttt{whisper-small} encoder followed by an average pooling and classification layers. For Slot Filling, the full \texttt{whisper-small} system is used with its tokenizer expanded to include the semantic tags in the training data.
For complete details of baseline models see \S~\ref{sec:appendix_baselines}.

\section{Results}
\label{sec:results}
We summarize the results of the shared tasks as well as contrast the general strategies used.  For descriptions of individual approaches, see~\S~\ref{sec:appendix_systems}.
\begin{table*}[t]
\centering
\small 
\setlength{\tabcolsep}{3pt}
\begin{tabular}{@{}lrrrcccccccc@{}}
\toprule
\textbf{Rank} & \textbf{Team} &  \textbf{WER} $\downarrow$ & \textbf{CER} $\downarrow$ &
\textbf{FT} & \textbf{LoRA} & \textbf{Add. data} &
\textbf{Aug.} & \textbf{Dialect-aware} & \textbf{LM} &
\textbf{Ensemble} & \textbf{Postproc.} \\
\midrule
1 & Wifaq              &  \textbf{43.24} & \textbf{19.75} & $\checkmark$ &    & $\checkmark$ &    &    & $\checkmark$ & $\checkmark$(4) &    \\
2 & Resonate           &  50.29 & 24.43 & $\checkmark$ &    &    & $\checkmark$ & $\checkmark$ &    &    & $\checkmark$ \\
3 & Thakaa              &  50.42 & 30.55 & $\checkmark$ & $\checkmark$ & $\checkmark$ &    & $\checkmark$ & $\checkmark$ & $\checkmark$(8) &    \\
4 & KAND CA    &  51.11 & 23.62 & $\checkmark$ &    & $\checkmark$ &    & $\checkmark$ &    &    & $\checkmark$ \\
5 & Ar\_India          &  52.91 & 25.36 & \cellcolor[gray]{0.8} &  \cellcolor[gray]{0.8}  & \cellcolor[gray]{0.8} & \cellcolor[gray]{0.8}   &  \cellcolor[gray]{0.8}  &  \cellcolor[gray]{0.8}  &  \cellcolor[gray]{0.8}  & \cellcolor[gray]{0.8}    \\
6 & Nile               &  56.90 & 27.67 & $\checkmark$ &    & $\checkmark$ & $\checkmark$ &    &    &    &    \\
7 & Itgan              &  57.24 & 28.05 &    & $\checkmark$ &    &    & $\checkmark$ &    &    & $\checkmark$ \\
8 & Sabaa              &  59.77 & 28.04 &    & $\checkmark$ &    &    & $\checkmark$ &    &    & $\checkmark$ \\
9 & Salesteq           &  64.20 & 35.80 & \cellcolor[gray]{0.8} & \cellcolor[gray]{0.8} & \cellcolor[gray]{0.8} & \cellcolor[gray]{0.8} & \cellcolor[gray]{0.8}& \cellcolor[gray]{0.8}& \cellcolor[gray]{0.8} & \cellcolor[gray]{0.8} \\
10 & Baseline &  81.37 & 45.89 & & & & & & & &  \\
\bottomrule
\end{tabular}%
\caption{Official Subtask~1.1 (Robust ASR) results and reported system characteristics,
ordered by country-average percentage WER and; percentage CER is
also shown, and lower is better. \textbf{FT}: task-specific fine-tuning
other than explicitly reported Low-Rank Adaptation (LoRA); \textbf{Add. data}: use of data beyond the official Subtask~1.1 data, excluding checkpoint
pretraining; \textbf{Aug.}: acoustic augmentation;
\textbf{Dialect-aware}: use of either separate adapters, conditioning, routing, or
dialect-specific normalization; \textbf{LM}: external language model decoding; \textbf{Ensemble}: combination of multiple ASR systems (with number of components); \textbf{Postproc.}: denotes output transformations. A checkmark indicates a reported final-system component. \textsc{ar\_india} and \textsc{Salesteq} did not submit system papers. }
\label{tab:nadi26-subtask11-systems}
\end{table*}

\subsection{Task 1: ASR}
\subsubsection{Subtask~1.1: Robust ASR}
Table~\ref{tab:nadi26-subtask11-systems} summarizes the nine leaderboard
entries. System descriptions were available for seven of the nine teams. Overall, the leaderboard does not identify a single winning recipe.
\textsc{Wifaq} (\(43.24\%\) WER, \(19.75\%\) CER) led \textsc{Resonate} by \(7.05\%\) absolute WER, whereas \textsc{Resonate} (\(50.29 \%\) WER, \(24.43 \%\) CER) and \textsc{Thakaa} (\(50.42 \%\) WER, \(30.55 \%\) CER) were
separated by only \(0.13\%\), although with a much larger gap in CER.

The top three represented distinct strategies: \textsc{Wifaq} combined four w2v-BERT~2.0~\cite{barrault2023seamless} CTC recognizers with five-gram KenLM~\cite{heafield2011kenlm} decoding and ROVER~\cite{fiscus1997post}; Resonate fully adapted one Arabic-specialized encoder-decoder model; and
\textsc{Thakaa} used a heterogeneous, dialect-aware ROVER ensemble. Interestingly, while \textsc{Wifaq} also dominated CER, \textsc{KAND CA} placed second on that metric, opting for a finetuned Whisper paired with a finetuned mT5 multidialect model~\cite{xue2021mt5} for spelling correction; the same strategy may explain its comparatively weaker WER, where it ranked fourth.
Teams utilized diverse supplementary training data, with \textsc{Wifaq} and \textsc{KAND CA} both including Casablanca~\cite{talafha2024casablanca} and \textsc{Thakaa} using the Subtask~1.3 data, whereas the second-ranked team, \textsc{Resonate}, avoided extra data completely.
Dialect-aware mechanisms, such as dialect-specific routing, likewise appeared at both higher and lower ranks. 




\begin{table*}[t]
\centering
\small
\renewcommand{\arraystretch}{1.05}
\tabcolsep5pt
\begin{tabular}{clcclcccccc}
\toprule

& &
\multicolumn{2}{c}{\textbf{Results}} &
\multicolumn{7}{c}{\textbf{System Features}} \\
\cmidrule(lr){3-4}
\cmidrule(lr){5-11}

\textbf{Rank} &
\textbf{Team} &
\textbf{WER} $\downarrow$ &
\textbf{CER} $\downarrow$ &
\textbf{Model} &
\textbf{Adapt.} &
\textbf{Add. Data} &
\textbf{Ens.} &
\textbf{Replay} &
\textbf{VAD} &
\textbf{Post.} \\
\midrule

1 &
\textbf{Salesteq} &
\textbf{44.22} &
18.05 &
\cellcolor[gray]{0.8} &
\cellcolor[gray]{0.8} &
\cellcolor[gray]{0.8} &
\cellcolor[gray]{0.8} &
\cellcolor[gray]{0.8} &
\cellcolor[gray]{0.8} &
\cellcolor[gray]{0.8} \\

2 &
Namaa &
44.65 &
\textbf{17.16} &
Cohere &
ZS &
  &
  &
  &
  &
  \\

3 &
Itgan &
46.71 &
18.78 &
Whisper &
LoRA &
$\checkmark$ &
$\checkmark$(5) &
  &
$\checkmark$ &
  \\

4 &
KAND CA &
48.07 &
17.28 &
Whisper &
FFT &
$\checkmark$ &
  &
$\checkmark$ &
  &
$\checkmark$ \\

5 &
Baseline &
57.91 &
23.66 &
Whisper &
  &
  &
  &
  &
  &
  \\
\bottomrule
\end{tabular}%

\caption{
Official results and main system features for Subtask~1.2
(Mixed-Dialect ASR) in percentage WER and CER.
\textbf{Adapt.}: Model adaption strategy including
zero-shot inference (ZS),
 low-rank adaptation (LoRA),
and full-model fine-tuning (FFT);
\textbf{Add. Data}: use of training data beyond the Task~1.2 development set;
\textbf{Ens.}: combination of multiple ASR systems (and number of components);
\textbf{Replay}: reuse of previously seen training examples during subsequent fine-tuning;
\textbf{VAD}: voice activity detection; and
\textbf{Post.}: explicit transcription post-processing.
A $\checkmark$ indicates that the corresponding component was used.
\textsc{Salesteq} did not submit a system paper.
 }
\label{tab:asr-1-2-res}

\end{table*}
\subsubsection{Subtask~1.2: Mixed-Dialect ASR}
As shown in Table~\ref{tab:asr-1-2-res}, \textsc{Salesteq} achieved the lowest WER at \(44.22\%\), closely followed by \textsc{Namaa Community} at \(44.65\%\). \textsc{Itgan} and \textsc{KAND CA} obtained WERs of \(46.71\%\) and \(48.07\%\), respectively. For CER, \textsc{NAMAA Community} achieved the best score (\(17.16\%\)), followed by \textsc{KAND CA} (\(17.28\%\)). 

The submitted systems followed different strategies: \textsc{NAMAA Community} used zero-shot inference with an Arabic-specific Cohere model~\cite{shaun_cassini_2026}, \textsc{Itgan} combined LoRA-adapted Whisper models using ROVER, while \textsc{KAND CA} relied on full-model fine-tuning with additional data and replay. The top ranked team, \textsc{Salesteq}, unfortunately did not submit a system paper.
For data, most submissions did not use external data for this task, with only \textsc{Itgan} using task 1.1 training data, and \textsc{KAND CA} using Casablanca~\cite{talafha2024casablanca} to supplement training.

\subsubsection{Subtask~1.3: Code-switched ASR} For code-switched ASR, the leaderboard, Table \ref{tab:nadi26-subtask13-systems}, is exceptionally tight at the top: the first three entries span
\(0.12\%\) absolute WER. 
Below that cluster, \textsc{KAND CA} trails the leader by \(0.8\%\) and \textsc{NAMAA Community} by \(2.57\%\), while \textsc{ar\_india} is a far outlier at \(73.76\%\). Hamza normalization lowers the top five teams' WER by roughly \(0.30\%\) and does not reorder the ranking. As in Subtask~1.1, the two metrics give different views: \textsc{Abjad AI} led on WER, whereas \textsc{Itgan} recorded the lowest CER of the entire leaderboard, and \textsc{Thakaa} ranked third by
WER but fourth by CER.  

One strategy that clearly paid off in this subtask was the multi-\textit{model} approach: all three top systems used some form of model combination (weight averaging or ROVER). As in Subtasks~1.1 and~1.2, supplementary training data approaches were diverse, with the \textsc{Abjad AI} team utilizing synthetic speech, and the third-ranked team utilizing TUNIFRA~\cite{choux2025tunifra} and Tunisian SLURP~\cite{slurptn}.

\subsection{Task~2: Spoken Dialect Identification} We provide the leaderboard for the SDID task in Table \ref{tab:adi-2-res}.
This task proved quite difficult compared to the in-domain validation performance, with the baseline model going from \(81.19\%\) accuracy and $C_{avg}$ of \(10.60\) to an accuracy of \(28.70\%\) and $C_{avg}$ of \(20.02\). However, the teams were able to make a large gain over this performance, with  \textsc{Thakaa} achieving the best accuracy at \(56.15\%\) and \textsc{Salesteq} achieving the best cost at \(8.64\). 

Multimodal approaches appeared to dominate performance, with
\textsc{Thakaa} using an ensemble mainly weighted towards a single acoustic model, Ara-BEST-RQ~\cite{elleuch2026ara}, and lightly weighted ($30\%$) text branches. A number of groups had reasonable success using Cohere's Arabic transcription model~\cite{shaun_cassini_2026}, with \textsc{AI Elites}, \textsc{KAND CA}, and \textsc{Abjad AI} all using it as their base model for finetuning, while \textsc{Thakaa} used it as part of their text-pipelines.


\subsection{Task~3: Text-to-Speech}
This task received a single submission from \textsc{KAND CA}~\cite{kand-nadi2026}, covering all \(4{,}001\) evaluation samples across the ten target dialects. As shown in Table~\ref{tab:tts-task3-results}, the system achieved aggregate scores of \(2.72\) UTMOS, \(3.57\) NISQA-MOS, and \(3.21\) NISQA-TTS, together with a WER of \(32.48\%\) and a CER of \(11.40\%\). Performance varied across dialects: Morocco obtained the highest UTMOS score (\(3.23\)), while Sudan achieved the highest NISQA-MOS (\(4.46\)) and NISQA-TTS (\(3.98\)) scores. Saudi Arabia showed the strongest intelligibility results, with the lowest WER (\(14.28\%\)) and CER (\(4.87\%\)).

\textsc{KAND CA} used \texttt{lahgtna-omnivoice-v2}, a 612M-parameter OmniVoice-based model~\cite{zhu2026omnivoice} for multi-dialect Arabic speech synthesis. Zero-shot voice cloning was applied using reference speech selected for each target dialect, with generated audio produced at 24~kHz. Arabic diacritics were retained to support pronunciation during synthesis. For dialects without a directly supported reference configuration, the closest available dialect was used: Saudi Arabic for UAE and Palestinian Arabic for Jordan.

Comparison against the baseline is instructive. Averaged across dialects, the \textsc{XTTS-v2} baseline outperforms the submission on all five metrics. This should be read with care. The baseline synthesizes all ten dialects from a single reference speaker with no dialect conditioning, and its perceptual scores are correspondingly near-invariant: UTMOS spans \(0.06\) across dialects and NISQA-MOS \(0.18\), against \(1.06\) and \(1.81\) for the submission. Metrics that barely register a system rendering every variety identically are largely insensitive to dialectal identity. UTMOSv2 and NISQA score general naturalness, while our WER and CER come from an ASR system carrying the same MSA bias this shared task exists to expose. Morocco is illustrative: the submission's highest UTMOS (\(3.23\), above baseline) coincides with its highest WER (\(55.46\%\) vs.\ \(27.34\%\)). We therefore take this result to reflect the limits of current automatic metrics rather than the futility of dialect-specific synthesis: informal listening confirms that the baseline renders every dialect in a largely uniform, MSA-leaning register that none of our metrics penalizes. Settling the matter properly requires human evaluation.

\subsection{Task~4: Spoken Language Translation}
As shown in Table~\ref{tab:ast-task4-results}, \textsc{Sense} achieved an overall BLEU score of \(14.19\) and a chrF score of \(34.11\). Dialect-level performance varied substantially, with the highest BLEU obtained for Jordan (\(27.46\)), followed by Palestine (\(20.32\)) and Egypt (\(19.84\)). Lower scores were observed for Morocco (\(8.51\)), Mauritania (\(6.84\)), and particularly the UAE (\(0.39\)). 

Only the \textsc{Abjad AI} team submitted a system paper. They directly finetuned  Whisper Large-v3 on the provided training data, with additional data augmentation.

\subsection{Task 5: Spoken Language Understanding}
\subsubsection{Subtask~5.1: Intent Recognition}
As shown in Table~\ref{tab:slu_5_1_res}, \textsc{Lynx} achieved the best performance, with a weighted F\textsubscript{1} of \(76.86\%\) and an accuracy of \(75.73\%\). \textsc{CIS-OpenSLU} ranked second with \(71.54\%\) weighted F\textsubscript{1} and \(72.50\%\) accuracy, followed by \textsc{A3da2 Messi} with \(69.92\%\) weighted F\textsubscript{1} and	\(70.07\%\) accuracy, and then \textsc{Aslema} with \(66.93\%\) weighted F\textsubscript{1} and \(66.13\%\) accuracy. \textsc{FARABI-Fraunhofer} ranked fifth, obtaining \(64.69\%\) weighted F\textsubscript{1} and \(65.93\%\) accuracy. The ranking is consistent across both metrics, with \textsc{Lynx} maintaining a clear lead over the remaining systems.
 
The top two groups and the fourth-ranked team submitted system papers, with \textsc{Lynx} and \textsc{CIS-OpenSLU} both using an ASR-to-TF-IDF feature pipeline, while \textsc{Aslema} utilized a LoRA finetuned multimodal LLM instead. Neither \textsc{Lynx} nor \textsc{CIS-OpenSLU} utilized external data, while \textsc{Aslema} generated additional synthetic data for training. A novel feature of \textsc{Lynx}'s approach was to filter unknown category data by training \texttt{MARBERTv2}~\cite{abdul2021arbert} on MASSIVE~\cite{fitzgerald2023massive} and conditioning on the probability sum over the valid categories.

\subsubsection{Subtask~5.2: Slot Filling}
As shown in Table~\ref{tab:slu-5-2-res}, \textsc{Aslema} achieved the best performance, with a CoER of \(59.53\%\) and a CVER of \(94.20\%\). \textsc{Zila} ranked second with a CoER of \(72.78\%\) and a CVER of \(100.36\%\).

Only \textsc{Aslema} provided a system paper, and their strategy was similar to their intent classification approach: LoRA finetuning \texttt{Qwen3-Omni-30B}~\cite{yang2025qwen3} on the SLURP-TN~\cite{slurptn} training split. Key to their strategy, however, was synthetic data generation, using \texttt{Gemini 3.6 Flash} and \texttt{3.1 Pro} to generate data, and \texttt{Gemini 3.6 Flash}, \texttt{3.1 Pro}, and \texttt{2.5 Pro} as a judging panel for data filtering.

\section{Discussion}
\label{sec:disc}
\paragraph{New model backbones.} One model appeared frequently through different submissions: Cohere's  \texttt{Transcribe} Arabic model~\cite{shaun_cassini_2026}. It accounted for two of the top performing systems in Subtask~1.1, the second place system in Subtask~1.2, the top system in Subtask~1.3, four of the described systems in Subtask~2, and the top performing system in Subtask~5.1. Aside from this model, we observe that the \textbf{Ara-BEST-RQ} self-supervised learning model~\cite{elleuch2026ara} proved a powerful component for the top Subtask~2 team's submission.

\paragraph{Multimodal SDID.} NADI 2025's SDID  task was dominated by end-to-end acoustic models~\cite{talafha-etal-2025-nadi}, yet this year we have seen a number of multimodal systems perform well (accounting for three of the top five systems), hinting at the importance of both textual and spoken features. Historically, ASR transcripts~\cite{malmasi2016arabic} as well as phone-based features~\cite{biadsy2009spoken} have been utilized in SDID, and multimodal systems may reflect an evolution of these approaches.

\paragraph{Ensembling.} Combined systems accounted for 10 of the 41 valid test submissions.\footnote{We received 48 in total; see Table~\ref{tab:participation} for the duplicate, unregistered, and invalid entries excluded} One real-world limitation of this approach, however, is the additional compute needed to both train and run inference of these models. To address low-compute scenarios, a potential direction for NADI would be to examine model efficiency across subtasks.

\paragraph{Is SDID Necessary?} A surprising finding from the Mixed-Dialect ASR task, was the lack of any groups integrating SDID into their pipeline. The growth of multi-dialect models which do not differentiate between dialects well, such as Cohere's Transcribe Arabic or Whisper, may lead teams to embrace a `one-size-fits all' approach to this task. Lack of data to train dialect-specific ASR models may contribute to this. Future work may benefit from directly comparing SDID-integrated pipelines and dialect-specific ASR with these more general approaches.

\paragraph{Lessons learned.} Reflecting on this year's shared task, we have identified a few areas to consider before the next edition of NADI. First, the TTS evaluation should be redesigned (we expand on this further in \S\ref{sec:limit}). Second, the SDID subtask may be too difficult. It could benefit from providing dialect coverage information as well as being restructured to evaluate  performance across several discrete domains, rather than as a purely zero-shot evaluation. Finally, an essential element of shared tasks is enabling teams to submit contrastive submissions. We believe we can achieve this by using evaluation platforms that better support display of variations in different systems.


\section{Conclusion}\label{sec:conclusion}

For NADI 2026, the second speech processing focused edition of NADI, we have expanded into more challenging subtasks (Robust, Mixed-dialect, and Code-switched ASR) as well as introduced new subtasks (TTS, SLT, SLU). This year we also enjoyed the joint-largest number of participating teams (21, tied with NADI 2022), and the second-largest number of system description papers (14), while expanding to eight subtasks. 

This year's shared task highlights a number of facets, including new pretrained Arabic ASR models with strong DA performance, the importance of multimodal approaches for SDID, and the dominance of ensemble-based approaches. In the future, we plan to examine performance across tasks under low-compute scenarios, more explicitly investigate the benefit of SDID systems as part of an ASR pipeline, and possibly expand the coverage to new dialects.

\section*{Limitations}\label{sec:limit}

\paragraph{Linguistic Coverage.} While our objective is to provide coverage of a wide number of Arabic dialects in this shared task, we acknowledge the gap between our coverage and the great diversity of Arabic dialects across the world. In comparison to last year's NADI shared task, we believe we have made slight progress in this regard by introducing tasks that include code-switching and sub-dialectal variation. In terms of numbers of dialects, the broadest coverage task, TTS, still only covers ten dialects compared to the 28 varieties identified by ISO.

\paragraph{Dual Use Implications.} While improving the spoken language processing methods for DA is essential in building equitable language technologies that match speaker's expectations for everyday language, these methods have potential for misuse, including for surveillance, or in the case of TTS, fraud or deception.

\paragraph{TTS Evaluation.} We acknowledge two limitations with our TTS subtask: low participation and lack of human evaluation. With respect to low participation, one team communicated an expectation that training data would be provided and withdrew on hearing this was not the case. Additionally, while other subtasks may have benefited from CodaBench's discoverability features, information about the TTS subtask was only disseminated through our website and emails. Likewise, lack of a submission platform may have inconvenienced participants, further discouraging participation. 

Next, while automatic, non-intrusive, metrics may capture some audio quality information and even some semantic information (via ASR WER), however, they lack a ground truth human evaluation to validate their performance and suitability across different Arabic dialects. 
While this year we focused on covering a wide number of dialects, an alternative would have been to provide lesser coverage of dialects, but in turn supplement the evaluation with human metrics.

These two factors limit the conclusions that can be drawn from this subtask, but does provides important organizational lessons which we hope will help address these issues in future iterations of NADI.

\paragraph{Orthographic Variability.} Dialectal Arabic has no standardized orthography, yet WER and CER score a hypothesis against a single reference, so a transcription a native speaker would accept can be counted as wrong for choosing a different valid spelling. Subtask~1.3 lets us bound this: collapsing hamza variants alone lowers the top five teams' WER by \(0.29\) points, about \(2\%\) of measured word error, without reordering the leaderboard. Rankings therefore appear robust, but absolute rates are less so: hamza is one axis among several, so \(2\%\) is a lower bound. Multi-reference WER~\cite{ali2015multi} was proposed for  this setting and could be adopted in future editions.

\section*{Acknowledgments}

Muhammad Abdul-Mageed acknowledges support from Canada Research Chairs (CRC), the Natural Sciences and Engineering Research Council of Canada (NSERC; RGPIN-2026-07098), the Social Sciences and Humanities Research Council of Canada (SSHRC; 895-2020-1004), Canadian Foundation for Innovation (CFI; 37771), Digital Research Alliance of Canada,\footnote{\href{https://alliancecan.ca}{https://alliancecan.ca}} and UBC ARC-Sockeye. The KFUPM authors gratefully acknowledge the support of the SDAIA–KFUPM Joint Research Center (JRC) for providing the computational resources that contributed to this work.

\normalem
\bibliography{custom, referece_from_nadi2024}

@inproceedings{abdul2020nadi,
  author       = {Muhammad Abdul{-}Mageed and
                  Chiyu Zhang and
                  Houda Bouamor and
                  Nizar Habash},
  editor       = {Imed Zitouni and
                  Muhammad Abdul{-}Mageed and
                  Houda Bouamor and
                  Fethi Bougares and
                  Mahmoud El{-}Haj and
                  Nadi Tomeh and
                  Wajdi Zaghouani},
  title        = {{NADI} 2020: The First Nuanced {Arabic} Dialect Identification Shared
                  Task},
  booktitle    = {Proceedings of the Fifth Arabic Natural Language Processing Workshop,
                  WANLP@COLING 2020, Barcelona, Spain (Online), December 12, 2020},
  pages        = {97--110},
  publisher    = {Association for Computational Linguistics},
  year         = {2020},
  url          = {https://www.aclweb.org/anthology/2020.wanlp-1.9/},
  bibsource    = {dblp computer science bibliography, https://dblp.org}
}

@inproceedings{abdul2021nadi,
  author       = {Muhammad Abdul{-}Mageed and
                  Chiyu Zhang and
                  AbdelRahim A. Elmadany and
                  Houda Bouamor and
                  Nizar Habash},
  editor       = {Nizar Habash and
                  Houda Bouamor and
                  Hazem M. Hajj and
                  Walid Magdy and
                  Wajdi Zaghouani and
                  Fethi Bougares and
                  Nadi Tomeh and
                  Ibrahim Abu Farha and
                  Samia Touileb},
  title        = {{NADI} 2021: The Second Nuanced {Arabic} Dialect Identification Shared
                  Task},
  booktitle    = {Proceedings of the Sixth Arabic Natural Language Processing Workshop,
                  {WANLP} 2021, Kyiv, Ukraine (Virtual), April 9, 2021},
  pages        = {244--259},
  publisher    = {Association for Computational Linguistics},
  year         = {2021},
  url          = {https://www.aclweb.org/anthology/2021.wanlp-1.28/},
  bibsource    = {dblp computer science bibliography, https://dblp.org}
}

@inproceedings{abdul2022nadi,
    title = "{NADI} 2022: The Third Nuanced {A}rabic Dialect Identification Shared Task",
    author = "Abdul-Mageed, Muhammad and Zhang, Chiyu and Elmadany, AbdelRahim and Bouamor, Houda and Habash, Nizar",
    booktitle = "Proceedings of the Seventh Arabic Natural Language Processing Workshop (WANLP)",
    month = dec, year = "2022", address = "Abu Dhabi, United Arab Emirates (Hybrid)",
    publisher = "Association for Computational Linguistics",
    url = "https://aclanthology.org/2022.wanlp-1.9/",
    doi = "10.18653/v1/2022.wanlp-1.9", pages = "85--97"
}

@inproceedings{abdul2023nadi,
    title = "{NADI} 2023: The Fourth Nuanced {A}rabic Dialect Identification Shared Task",
    author = "Abdul-Mageed, Muhammad and Elmadany, AbdelRahim and Zhang, Chiyu and Nagoudi, El Moatez Billah and Bouamor, Houda and Habash, Nizar",
    booktitle = "Proceedings of ArabicNLP 2023",
    month = dec, year = "2023", address = "Singapore (Hybrid)",
    publisher = "Association for Computational Linguistics",
    url = "https://aclanthology.org/2023.arabicnlp-1.62/",
    doi = "10.18653/v1/2023.arabicnlp-1.62", pages = "600--613"
}

@inproceedings{abdul2024nadi,
    title = "{NADI} 2024: The Fifth Nuanced {A}rabic Dialect Identification Shared Task",
    author = "Abdul-Mageed, Muhammad and Keleg, Amr and Elmadany, AbdelRahim and Zhang, Chiyu and Hamed, Injy and Magdy, Walid and Bouamor, Houda and Habash, Nizar",
    booktitle = "Proceedings of the Second Arabic Natural Language Processing Conference",
    month = aug, year = "2024", address = "Bangkok, Thailand",
    publisher = "Association for Computational Linguistics",
    url = "https://aclanthology.org/2024.arabicnlp-1.79/",
    doi = "10.18653/v1/2024.arabicnlp-1.79", pages = "709--728"
}

@inproceedings{talafha2024casablanca,
    title = "{C}asablanca: Data and Models for Multidialectal {A}rabic Speech Recognition",
    author = "Talafha, Bashar and Kadaoui, Karima and Magdy, Samar Mohamed and Habiboullah, Mariem and Chafei, Chafei Mohamed and El-Shangiti, Ahmed Oumar and Zayed, Hiba and Tourad, Mohamedou Cheikh and Alhamouri, Rahaf and Assi, Rwaa and Alraeesi, Aisha and Mohamed, Hour and Alwajih, Fakhraddin and Mohamed, Abdelrahman and El Mekki, Abdellah and Nagoudi, El Moatez Billah and Saadia, Benelhadj Djelloul Mama and Alsayadi, Hamzah A. and Al-Dhabyani, Walid and Shatnawi, Sara and Ech-chammakhy, Yasir and Makouar, Amal and Berrachedi, Yousra and Jarrar, Mustafa and Shehata, Shady and Berrada, Ismail and Abdul-Mageed, Muhammad",
    booktitle = "Proceedings of the 2024 Conference on Empirical Methods in Natural Language Processing",
    month = nov, year = "2024", address = "Miami, Florida, USA",
    publisher = "Association for Computational Linguistics",
    url = "https://aclanthology.org/2024.emnlp-main.1211/",
    doi = "10.18653/v1/2024.emnlp-main.1211", pages = "21745--21758"
}

@inproceedings{elleuch25_interspeech,
  title     = {{ADI-20: Arabic Dialect Identification dataset and models}},
  author    = {Haroun Elleuch and Salima Mdhaffar and Yannick Estève and Fethi Bougares},
  year      = {{2025}},
  booktitle = {{Interspeech 2025}},
  pages     = {{2775--2779}},
  doi       = {{10.21437/Interspeech.2025-884}},
  issn      = {{2958-1796}},
}

@inproceedings{toyin2023artst,
    title = "{A}r{TST}: {A}rabic Text and Speech Transformer",
    author = "Toyin, Hawau Olamide and Djanibekov, Amirbek and Kulkarni, Ajinkya and Aldarmaki, Hanan",
    booktitle = "Proceedings of ArabicNLP 2023",
    month = dec, year = "2023", address = "Singapore (Hybrid)",
    publisher = "Association for Computational Linguistics",
    url = "https://aclanthology.org/2023.arabicnlp-1.5/",
    doi = "10.18653/v1/2023.arabicnlp-1.5", pages = "41--51"
}

@inproceedings{madar-2018-bouamor,
  author       = {Houda Bouamor and
                  Nizar Habash and
                  Mohammad Salameh and
                  Wajdi Zaghouani and
                  Owen Rambow and
                  Dana Abdulrahim and
                  Ossama Obeid and
                  Salam Khalifa and
                  Fadhl Eryani and
                  Alexander Erdmann and
                  Kemal Oflazer},
  editor       = {Nicoletta Calzolari and
                  Khalid Choukri and
                  Christopher Cieri and
                  Thierry Declerck and
                  Sara Goggi and
                  K{\^{o}}iti Hasida and
                  Hitoshi Isahara and
                  Bente Maegaard and
                  Joseph Mariani and
                  H{\'{e}}l{\`{e}}ne Mazo and
                  Asunci{\'{o}}n Moreno and
                  Jan Odijk and
                  Stelios Piperidis and
                  Takenobu Tokunaga},
  title        = {The {MADAR} {A}rabic Dialect Corpus and Lexicon},
  booktitle    = {Proceedings of the Eleventh International Conference on Language Resources
                  and Evaluation, {LREC} 2018, Miyazaki, Japan, May 7-12, 2018},
  publisher    = {European Language Resources Association {(ELRA)}},
  year         = {2018},
  url          = {http://www.lrec-conf.org/proceedings/lrec2018/summaries/351.html},
  bibsource    = {dblp computer science bibliography, https://dblp.org}
}

@misc{lee2022nist,
  title={{NIST} 2022 language recognition evaluation plan},
  author={Lee, Yooyoung and Greenberg, Craig and Mason, Lisa and Singer, Elliot},
  year={2022},
  howpublished={National Institute of Standards and Technology},
  url={https://www.nist.gov/publications/nist-2022-language-recognition-evaluation-plan}
}

@inproceedings{valk2021voxlingua107,
  title={VoxLingua107: a dataset for spoken language recognition},
  author={Valk, J{\"o}rgen and Alum{\"a}e, Tanel},
  booktitle={2021 IEEE Spoken Language Technology Workshop (SLT)},
  pages={652--658},
  year={2021},
  organization={IEEE}
}

@inproceedings{baas23_interspeech,
  title     = {Voice Conversion With Just Nearest Neighbors},
  author    = {Matthew Baas and Benjamin {van Niekerk} and Herman Kamper},
  year      = {2023},
  booktitle = {Interspeech 2023},
  pages     = {2053--2057},
  doi       = {10.21437/Interspeech.2023-419},
  issn      = {2958-1796},
}

@inproceedings{sullivan23_interspeech,
  title     = {On the Robustness of {A}rabic Speech Dialect Identification},
  author    = {Peter Sullivan and AbdelRahim Elmadany and Muhammad Abdul-Mageed},
  year      = {2023},
  booktitle = {Interspeech 2023},
  pages     = {5326--5330},
  doi       = {10.21437/Interspeech.2023-1005},
  issn      = {2958-1796},
}

@inproceedings{ali2017speech,
  title={Speech recognition challenge in the wild: {A}rabic {MGB-3}},
  author={Ali, Ahmed and Vogel, Stephan and Renals, Steve},
  booktitle={2017 IEEE Automatic Speech Recognition and Understanding Workshop (ASRU)},
  pages={316--322},
  year={2017},
  organization={IEEE}
}

@inproceedings{ali2019mgb,
  title={The {MGB}-5 challenge: Recognition and dialect identification of dialectal {A}rabic speech},
  author={Ali, Ahmed and Shon, Suwon and Samih, Younes and Mubarak, Hamdy and Abdelali, Ahmed and Glass, James and Renals, Steve and Choukri, Khalid},
  booktitle={2019 IEEE Automatic Speech Recognition and Understanding Workshop (ASRU)},
  pages={1026--1033},
  year={2019},
  organization={IEEE}
}

@inproceedings{gulati2020conformer,
  title     = {{Conformer: Convolution-augmented Transformer for Speech Recognition}},
  author    = {Anmol Gulati and James Qin and Chung-Cheng Chiu and Niki Parmar and Yu Zhang and Jiahui Yu and Wei Han and Shibo Wang and Zhengdong Zhang and Yonghui Wu and Ruoming Pang},
  year      = {2020},
  booktitle = {{Interspeech 2020}},
  pages     = {5036--5040},
  doi       = {10.21437/Interspeech.2020-3015},
  issn      = {2958-1796},
}

@inproceedings{graves2006connectionist,
  title={Connectionist temporal classification: labelling unsegmented sequence data with recurrent neural networks},
  author={Graves, Alex and Fern{\'a}ndez, Santiago and Gomez, Faustino and Schmidhuber, J{\"u}rgen},
  booktitle={Proceedings of the 23rd international conference on Machine learning},
  pages={369--376},
  year={2006}
}

@article{barrault2023seamless,
  title={Seamless: Multilingual Expressive and Streaming Speech Translation},
  author={Barrault, Lo{\"\i}c and Chung, Yu-An and Meglioli, Mariano Coria and Dale, David and Dong, Ning and Duppenthaler, Mark and Duquenne, Paul-Ambroise and Ellis, Brian and Elsahar, Hady and Haaheim, Justin and others},
  journal={arXiv preprint arXiv:2312.05187},
  year={2023},
  url = {https://arxiv.org/abs/2312.05187},
}

@inproceedings{ardila-etal-2020-common,
    title = "Common {V}oice: A Massively-Multilingual Speech Corpus",
    author = "Ardila, Rosana and Branson, Megan and Davis, Kelly and Kohler, Michael and Meyer, Josh and Henretty, Michael and Morais, Reuben and Saunders, Lindsay and Tyers, Francis and Weber, Gregor",
    booktitle = "Proceedings of the Twelfth Language Resources and Evaluation Conference",
    month = may, year = "2020", address = "Marseille, France",
    publisher = "European Language Resources Association",
    url = "https://aclanthology.org/2020.lrec-1.520/",
    pages = "4218--4222", isbn = "979-10-95546-34-4"
}

@inproceedings{al-fetyani2022masc,
  title        = {{MASC}: Massive {Arabic} Speech Corpus},
  author       = {Al-Fetyani, Mohammad and Al-Barham, Muhammad and Abandah, Gheith A. and Alsharkawi, Adham and Dawas, Maha},
  booktitle    = {Proceedings of the 2022 IEEE Spoken Language Technology Workshop (SLT)},
  year         = {2022},
  pages        = {1002206},
  doi          = {10.1109/SLT54892.2023.10022652},
}

@inproceedings{talafha-etal-2025-nadi,
    title = "{NADI} 2025: The First Multidialectal {A}rabic Speech Processing Shared Task",
    author = "Talafha, Bashar  and
      Toyin, Hawau Olamide  and
      Sullivan, Peter  and
      Elmadany, AbdelRahim A.  and
      Juma, Abdurrahman  and
      Djanibekov, Amirbek  and
      Zhang, Chiyu  and
      Alshehhi, Hamad  and
      Aldarmaki, Hanan  and
      Jarrar, Mustafa  and
      Habash, Nizar  and
      Abdul-Mageed, Muhammad",
    editor = "Darwish, Kareem  and
      Ali, Ahmed  and
      Abu Farha, Ibrahim  and
      Touileb, Samia  and
      Zitouni, Imed  and
      Abdelali, Ahmed  and
      Al-Ghamdi, Sharefah  and
      Alkhereyf, Sakhar  and
      Zaghouani, Wajdi  and
      Khalifa, Salam  and
      AlKhamissi, Badr  and
      Almatham, Rawan  and
      Hamed, Injy  and
      Alyafeai, Zaid  and
      Alowisheq, Areeb  and
      Inoue, Go  and
      Mrini, Khalil  and
      Alshammari, Waad",
    booktitle = "Proceedings of The Third Arabic Natural Language Processing Conference: Shared Tasks",
    month = nov,
    year = "2025",
    address = "Suzhou, China",
    publisher = "Association for Computational Linguistics",
    url = "https://aclanthology.org/2025.arabicnlp-sharedtasks.99/",
    doi = "10.18653/v1/2025.arabicnlp-sharedtasks.99",
    pages = "720--733",
    ISBN = "979-8-89176-356-2",
}

@inproceedings{abdullah25_interspeech,
  title     = {{Voice Conversion Improves Cross-Domain Robustness  for Spoken {Arabic} Dialect Identification}},
  author    = {Badr M. Abdullah and Matthew Baas and Bernd Möbius and Dietrich Klakow},
  year      = {2025},
  booktitle = {{Interspeech 2025}},
  pages     = {2790--2794},
  doi       = {10.21437/Interspeech.2025-1809},
  issn      = {2958-1796},
}

@misc{tan2021surveyneuralspeechsynthesis,
      title={A Survey on Neural Speech Synthesis}, 
      author={Xu Tan and Tao Qin and Frank Soong and Tie-Yan Liu},
      year={2021},
      eprint={2106.15561},
      archivePrefix={arXiv},
      primaryClass={eess.AS},
      url={https://arxiv.org/abs/2106.15561}, 
}

@misc{googlecloud_gemini_tts,
  author       = {{Google Cloud}},
  title        = {{Gemini-TTS} | {Cloud Text-to-Speech}},
  howpublished = {\url{https://cloud.google.com/text-to-speech/docs/gemini-tts}},
  year         = {2026},
  note         = {Accessed: Aug. 9, 2026}
}

@misc{xiang2026qwenaudio30ttsfreelycontrollablehighly,
      title={{Qwen-Audio-3.0-TTS}: Freely Controllable and Highly Robust Speech Synthesis with Multi-Stage Training Paradigm}, 
      author={Bajian Xiang and Cheng Wen and Han Zhao and Hao Wang and Haoxu Wang and Jiawei Jin and Jiayan Cui and Jie Chen and Mengxi Nie and Tianyu Zhao and Weiqin Li and Xiang Lv and Xiangang Li and Yang Xiang and Yang Zhou},
      year={2026},
      eprint={2607.23938},
      archivePrefix={arXiv},
      primaryClass={eess.AS},
      url={https://arxiv.org/abs/2607.23938}, 
}

@misc{casanova2024xttsmassivelymultilingualzeroshot,
      title={{XTTS}: a Massively Multilingual Zero-Shot Text-to-Speech Model}, 
      author={Edresson Casanova and Kelly Davis and Eren Gölge and Görkem Göknar and Iulian Gulea and Logan Hart and Aya Aljafari and Joshua Meyer and Reuben Morais and Samuel Olayemi and Julian Weber},
      year={2024},
      eprint={2406.04904},
      archivePrefix={arXiv},
      primaryClass={eess.AS},
      url={https://arxiv.org/abs/2406.04904}, 
}

@misc{ao2022speecht5unifiedmodalencoderdecoderpretraining,
      title={{SpeechT5}: Unified-Modal Encoder-Decoder Pre-Training for Spoken Language Processing}, 
      author={Junyi Ao and Rui Wang and Long Zhou and Chengyi Wang and Shuo Ren and Yu Wu and Shujie Liu and Tom Ko and Qing Li and Yu Zhang and Zhihua Wei and Yao Qian and Jinyu Li and Furu Wei},
      year={2022},
      eprint={2110.07205},
      archivePrefix={arXiv},
      primaryClass={eess.AS},
      url={https://arxiv.org/abs/2110.07205}, 
}

@misc{elevenlabs_multilingual_v2,
  author       = {{ElevenLabs}},
  title        = {Eleven {Multilingual} v2},
  howpublished = {\url{https://elevenlabs.io/blog/eleven-multilingual-v2}},
  year         = {2023},
  note         = {Accessed: Aug. 9, 2026}
}

@misc{silma_arabic_tts,
  author       = {{SILMA AI}},
  title        = {Arabic {Text-to-Speech} ({TTS}) Models},
  howpublished = {\url{https://silma.ai/arabic-tts-models}},
  year         = {2026},
  note         = {Accessed: Aug. 9, 2026}
}

@misc{smolarek2019amazonpolly,
  author       = {Smolarek, Marta},
  title        = {Amazon {Polly} adds {Arabic} language support},
  howpublished = {AWS Machine Learning Blog},
  url          = {https://aws.amazon.com/blogs/machine-learning/amazon-polly-adds-arabic-language-support/},
  year         = {2019},
  month        = apr,
  note         = {Accessed: Aug. 9, 2026}
}

@inproceedings{Khamis_2026,
   title={{LLM-to-Speech}: A Synthetic Data Pipeline for Training Dialectal Text-to-Speech Models},
   url={http://dx.doi.org/10.18653/v1/2026.abjadnlp-1.6},
   DOI={10.18653/v1/2026.abjadnlp-1.6},
   booktitle={Proceedings of the 2nd Workshop on NLP for Languages Using Arabic Script},
   publisher={Association for Computational Linguistics},
   author={Khamis, Ahmed and Ahmed, Hesham Ali},
   year={2026},
   pages={47–54} }

@misc{OmarSamir,
      author = {Omar Samir and Youssef Waleed and Youssef Tamer and Amir Mohamed},
      title = {Fine-Tuning {XTTS V2} for {E}gyptian {Arabic}},
      year = {2024},
      url = {https://github.com/joejoe03/Egyptian-Text-To-Speech},
}

@misc{chen2026habibilayingopensourcefoundation,
      title={Habibi: Laying the Open-Source Foundation of Unified-Dialectal {Arabic} Speech Synthesis}, 
      author={Yushen Chen and Junzhe Liu and Yujie Tu and Zhikang Niu and Yuzhe Liang and Chunyu Qiang and Chen Zhang and Kai Yu and Xie Chen},
      year={2026},
      eprint={2601.13802},
      archivePrefix={arXiv},
      primaryClass={cs.CL},
      url={https://arxiv.org/abs/2601.13802}, 
}

@misc{oddadmix_lahgtna_chatterbox,
  author       = {{oddadmix}},
  title        = {lahgtna-chatterbox-v1},
  howpublished = {Hugging Face},
  url          = {https://huggingface.co/oddadmix/lahgtna-chatterbox-v1},
  year         = {2024},
  note         = {Accessed: Aug. 9, 2026}
}

@inproceedings{baba2024utmosv2,
  title     = {The {T05} System for the {V}oice{MOS} {C}hallenge 2024: Transfer Learning from Deep Image Classifier to Naturalness {MOS} Prediction of High-Quality Synthetic Speech},
  author    = {Baba, Kaito and Nakata, Wataru and Saito, Yuki and Saruwatari, Hiroshi},
  booktitle = {IEEE Spoken Language Technology Workshop (SLT)},
  year      = {2024},
  pages     = {818--824},
  doi       = {10.1109/SLT61566.2024.10832315},
}

@inproceedings{Mittag_2021, 
   title={NISQA: A Deep CNN-Self-Attention Model for Multidimensional Speech Quality Prediction with Crowdsourced Datasets},
   url={http://dx.doi.org/10.21437/Interspeech.2021-299},
   DOI={10.21437/interspeech.2021-299},
   booktitle={Interspeech 2021},
   publisher={ISCA},
   author={Mittag, Gabriel and Naderi, Babak and Chehadi, Assmaa and Möller, Sebastian},
   year={2021},
   month=Aug, pages={2127–2131},
}

@inproceedings{mittag20_interspeech,
  title     = {{Deep Learning Based Assessment of Synthetic Speech Naturalness}},
  author    = {Gabriel Mittag and Sebastian Möller},
  year      = {2020},
  booktitle = {{Interspeech 2020}},
  pages     = {1748--1752},
  doi       = {10.21437/Interspeech.2020-2382},
  issn      = {2958-1796},
}

@article{rabiner1989tutorial,
  author  = {Rabiner, Lawrence R.},
  title   = {A Tutorial on Hidden {M}arkov Models and Selected Applications in Speech Recognition},
  journal = {Proceedings of the IEEE},
  volume  = {77},
  number  = {2},
  pages   = {257--286},
  year    = {1989},
  publisher = {IEEE}
}

@article{hinton2012deep,
  author  = {Hinton, Geoffrey and Deng, Li and Yu, Dong and Dahl, George E. and Mohamed, Abdel-rahman and Jaitly, Navdeep and Senior, Andrew and Vanhoucke, Vincent and Nguyen, Patrick and Sainath, Tara N. and Kingsbury, Brian},
  title   = {Deep Neural Networks for Acoustic Modeling in Speech Recognition: The Shared Views of Four Research Groups},
  journal = {IEEE Signal Processing Magazine},
  volume  = {29},
  number  = {6},
  pages   = {82--97},
  year    = {2012}
}

@inproceedings{chan2016listen,
  author    = {Chan, William and Jaitly, Navdeep and Le, Quoc and Vinyals, Oriol},
  title     = {Listen, Attend and Spell: A Neural Network for Large Vocabulary Conversational Speech Recognition},
  booktitle = {Proceedings of ICASSP},
  pages     = {4960--4964},
  year      = {2016}
}

@inproceedings{baevski2020wav2vec,
  author    = {Baevski, Alexei and Zhou, Henry and Mohamed, Abdelrahman and Auli, Michael},
  title     = {wav2vec 2.0: A Framework for Self-Supervised Learning of Speech Representations},
  booktitle = {Advances in Neural Information Processing Systems (NeurIPS)},
  volume    = {33},
  pages     = {12449--12460},
  year      = {2020}
}

@article{hsu2021hubert,
  author  = {Hsu, Wei-Ning and Bolte, Benjamin and Tsai, Yao-Hung Hubert and Lakhotia, Kushal and Salakhutdinov, Ruslan and Mohamed, Abdelrahman},
  title   = {{HuBERT}: Self-Supervised Speech Representation Learning by Masked Prediction of Hidden Units},
  journal = {IEEE/ACM Transactions on Audio, Speech, and Language Processing},
  volume  = {29},
  pages   = {3451--3460},
  year    = {2021}
}

@article{pratap2024scaling,
  title={Scaling speech technology to 1,000+ languages},
  author={Pratap, Vineel and Tjandra, Andros and Shi, Bowen and Tomasello, Paden and Babu, Arun and Kundu, Sayani and Elkahky, Ali and Ni, Zhaoheng and Vyas, Apoorv and Fazel-Zarandi, Maryam and others},
  journal={Journal of Machine Learning Research},
  volume={25},
  number={97},
  pages={1--52},
  year={2024}
}

@inproceedings{ozyilmaz2025overcoming,
  title     = {{Overcoming Data Scarcity in Multi-Dialectal Arabic ASR via Whisper Fine-Tuning}},
  author    = {Ömer Tarik Özyilmaz and Matt Coler and Matias Valdenegro-Toro},
  year      = {2025},
  booktitle = {{Interspeech 2025}},
  pages     = {1158--1162},
  doi       = {10.21437/Interspeech.2025-2260},
  issn      = {2958-1796},
}

@inproceedings{wang2024open,
  title     = {{Open Universal Arabic ASR Leaderboard}},
  author    = {Yingzhi Wang and Anas Alhmoud and Muhammad Alqurishi},
  year      = {2025},
  booktitle = {{Interspeech 2025}},
  pages     = {1168--1172},
  doi       = {10.21437/Interspeech.2025-184},
  issn      = {2958-1796},
}

@inproceedings{kopikar2026amchi,
  title={Amchi-Low Resource Language {ASR} with Crowdsourced Post-Processing},
  author={Kopikar, Moksh and Mandloi, Naman},
  booktitle={Proc. SpeechProsody 2026},
  pages={924--925},
  year={2026}
}

@inproceedings{zhuo2025vietasr,
  title     = {{VietASR: Achieving Industry-level Vietnamese ASR with 50-hour labeled data and Large-Scale Speech Pretraining}},
  author    = {Jianheng Zhuo and Yifan Yang and Yiwen Shao and Yong Xu and Dong Yu and Kai Yu and Xie Chen},
  year      = {2025},
  booktitle = {{Interspeech 2025}},
  pages     = {1163--1167},
  doi       = {10.21437/Interspeech.2025-398},
  issn      = {2958-1796},
}

@inproceedings{li2025context,
  title     = {{In-context Language Learning for Endangered Languages in Speech Recognition}},
  author    = {Zhaolin Li and Jan Niehues},
  year      = {2025},
  booktitle = {{Interspeech 2025}},
  pages     = {738--742},
  doi       = {10.21437/Interspeech.2025-2626},
  issn      = {2958-1796},
}

@article{alabi2024afrihubert,
  title={{AfriHuBERT}: A self-supervised speech representation model for {A}frican languages},
  author={Alabi, Jesujoba O and Liu, Xuechen and Klakow, Dietrich and Yamagishi, Junichi},
  journal={arXiv preprint arXiv:2409.20201},
  year={2024},
  url = {https://arxiv.org/abs/2409.20201},
}

@inproceedings{blaschke2025multi,
  title     = {{A Multi-Dialectal Dataset for German Dialect ASR and Dialect-to-Standard Speech Translation}},
  author    = {Verena Blaschke and Miriam Winkler and Constantin Förster and Gabriele Wenger-Glemser and Barbara Plank},
  year      = {2025},
  booktitle = {{Interspeech 2025}},
  pages     = {913--917},
  doi       = {10.21437/Interspeech.2025-318},
  issn      = {2958-1796},
}

@inproceedings{xu2025leveraging,
  title     = {{Leveraging LLM and Self-Supervised Training Models for Speech Recognition in Chinese Dialects: A Comparative Analysis}},
  author    = {Tianyi Xu and Hongjie Chen and Qing Wang and Lv Hang and Jian Kang and Jie Li and Zhennan Lin and Yongxiang Li and Lei Xie},
  year      = {2025},
  booktitle = {{Interspeech 2025}},
  pages     = {584--588},
  doi       = {10.21437/Interspeech.2025-1669},
  issn      = {2958-1796},
}

@article{serditova2025automatic,
  title={Automatic speech recognition biases in {N}ewcastle {E}nglish: An error analysis},
  author={Serditova, Dana and Tang, Kevin and Steffens, Jochen},
  journal={arXiv preprint arXiv:2506.16558},
  year={2025},
  url = {https://arxiv.org/abs/2506.16558},
}

@inproceedings{mojarad2025automatic,
  title     = {{Automatic Speech Recognition of African American English: Lexical and Contextual Effects}},
  author    = {Hamid Mojarad and Kevin Tang},
  year      = {2025},
  booktitle = {{Interspeech 2025}},
  pages     = {3883--3887},
  doi       = {10.21437/Interspeech.2025-1511},
  issn      = {2958-1796},
}

@inproceedings{mehralian2025leveraging,
  title={Leveraging Geographic Metadata for Dialect-Aware Speech Recognition},
  author={Mehralian, Pouya and {Van hamme}, Hugo},
  booktitle={Interspeech 2025},
  pages={1153--1157},
  year={2025},
  doi={10.21437/Interspeech.2025-1839}
}

@inproceedings{bagat2025mixture,
  title     = {{Mixture of LoRA Experts for Low-Resourced Multi-Accent Automatic Speech Recognition}},
  author    = {Raphaël Bagat and Irina Illina and Emmanuel Vincent},
  year      = {2025},
  booktitle = {{Interspeech 2025}},
  pages     = {1143--1147},
  doi       = {10.21437/Interspeech.2025-775},
  issn      = {2958-1796},
}

@inproceedings{kumar2026jointly,
  title     = {{Jointly Improving Dialect Identification and ASR in Indian Languages using Multimodal Feature Fusion}},
  author    = {Saurabh Kumar and  Amartyaveer and Prasanta Kumar Ghosh},
  year      = {2025},
  booktitle = {{Interspeech 2025}},
  pages     = {2770--2774},
  doi       = {10.21437/Interspeech.2025-421},
  issn      = {2958-1796},
}

@inproceedings{biswas2025adapting,
  title={Adapting {W}hisper for low-resource {Hindi-English} code-mix speech with on-the-fly augmentation \& {LLM}-synthesised data},
  author={Biswas, Astik and Shevelev, Oleg and Abdaoui, Amine and Tyagi, Vivek and Boumadane, Abdelmoumene},
  booktitle={Interspeech 2025},
  pages={4293--4297},
  year={2025},
  doi={10.21437/Interspeech.2025-447}
}

@inproceedings{yang2025adapting,
  title     = {{Adapting Whisper for Parameter-efficient Code-Switching Speech Recognition via Soft Prompt Tuning}},
  author    = {Hongli Yang and Yizhou Peng and Hao Huang and Sheng Li},
  year      = {2025},
  booktitle = {{Interspeech 2025}},
  pages     = {5203--5207},
  doi       = {10.21437/Interspeech.2025-2549},
  issn      = {2958-1796},
}

@inproceedings{gong2025br,
  title     = {{BR-ASR: Efficient and Scalable Bias Retrieval Framework for Contextual Biasing ASR in Speech LLM}},
  author    = {Xun Gong and Anqi Lv and Wangyou Zhang and Zhiming Wang and Huijia Zhu and Yanmin Qian},
  year      = {2025},
  booktitle = {{Interspeech 2025}},
  pages     = {4043--4047},
  doi       = {10.21437/Interspeech.2025-326},
  issn      = {2958-1796},
}

@inproceedings{wang2025rag,
  title     = {{RAG-Boost: Retrieval-Augmented Generation Enhanced LLM-based Speech Recognition}},
  author    = {Pengcheng Wang and Sheng Li and Takahiro Shinozaki},
  year      = {2025},
  booktitle = {{Workshop on Multilingual Conversational Speech Language Model (MLC-SLM)}},
  pages     = {54--55},
}

@inproceedings{prakash2025better,
  title     = {{Better Pseudo-labeling with Multi-ASR Fusion and Error Correction by SpeechLLM}},
  author    = {Jeena Prakash and Blessingh Kumar and Kadri Hacioglu and Bidisha Sharma and Sindhuja Gopalan and Malolan Chetlur and Shankar Venkatesan and Andreas Stolcke},
  year      = {2025},
  booktitle = {{Interspeech 2025}},
  pages     = {579--583},
  doi       = {10.21437/Interspeech.2025-1707},
  issn      = {2958-1796},
}

@inproceedings{desplanques20_interspeech,
  title     = {{ECAPA-TDNN: Emphasized Channel Attention, Propagation and Aggregation in TDNN Based Speaker Verification}},
  author    = {Brecht Desplanques and Jenthe Thienpondt and Kris Demuynck},
  year      = {2020},
  booktitle = {{Interspeech 2020}},
  pages     = {3830--3834},
  doi       = {10.21437/Interspeech.2020-2650},
  issn      = {2958-1796},
}

@inproceedings{dehak2011language,
  title={Language recognition via i-vectors and dimensionality reduction},
  author={Dehak, Najim and Torres-Carrasquillo, Pedro A and Reynolds, Douglas and Dehak, Reda},
  booktitle={Interspeech 2011},
  pages={857--860},
  year={2011},
  doi={10.21437/Interspeech.2011-328}
}

@inproceedings{snyder2018spoken,
  title={Spoken language recognition using x-vectors.},
  author={Snyder, David and Garcia-Romero, Daniel and McCree, Alan and Sell, Gregory and Povey, Daniel and Khudanpur, Sanjeev},
  booktitle={Odyssey},
  volume={2018},
  pages={105--111},
  year={2018}
}

@inproceedings{babu22_interspeech,
  title     = {{XLS-R: Self-supervised Cross-lingual Speech Representation Learning at Scale}},
  author    = {Arun Babu and Changhan Wang and Andros Tjandra and Kushal Lakhotia and Qiantong Xu and Naman Goyal and Kritika Singh and Patrick {von Platen} and Yatharth Saraf and Juan Pino and Alexei Baevski and Alexis Conneau and Michael Auli},
  year      = {2022},
  booktitle = {{Interspeech 2022}},
  pages     = {2278--2282},
  doi       = {10.21437/Interspeech.2022-143},
  issn      = {2958-1796},
}

@article{vaswani2017attention,
  title={Attention is all you need},
  author={Vaswani, Ashish and Shazeer, Noam and Parmar, Niki and Uszkoreit, Jakob and Jones, Llion and Gomez, Aidan N and Kaiser, 
 
  Lukasz and Polosukhin, Illia},
  journal={Advances in neural information processing systems},
  volume={30},
  year={2017}
}

@inproceedings{kang2022deep,
    title = "Deep learning-based end-to-end spoken language identification system for domain-mismatched scenario",
    author = "Kang, Woohyun and Alam, Md Jahangir and Fathan, Abderrahim",
    booktitle = "Proceedings of the Thirteenth Language Resources and Evaluation Conference",
    month = jun, year = "2022", address = "Marseille, France",
    publisher = "European Language Resources Association",
    url = "https://aclanthology.org/2022.lrec-1.798/", pages = "7339--7343"
}

@article{dey2023cross,
  title={Cross-corpora spoken language identification with domain diversification and generalization},
  author={Dey, Spandan and Sahidullah, Md and Saha, Goutam},
  journal={Computer Speech \& Language},
  volume={81},
  pages={101489},
  year={2023},
  publisher={Elsevier}
}

@inproceedings{shon2017qcri,
  title={{MIT-QCRI} {Arabic} dialect identification system for the 2017 multi-genre broadcast challenge},
  author={Shon, Suwon and Ali, Ahmed and Glass, James},
  booktitle={2017 IEEE Automatic Speech Recognition and Understanding Workshop (ASRU)},
  pages={374--380},
  year={2017},
  organization={IEEE}
}

@inproceedings{elleuch2025elyadata,
    title = "{ELYADATA} {\&} {LIA} at {NADI} 2025: {ASR} and {ADI} Subtasks",
    author = {Elleuch, Haroun and Saidi, Youssef and Mdhaffar, Salima and Est{\`e}ve, Yannick and Bougares, Fethi},
    booktitle = "Proceedings of The Third Arabic Natural Language Processing Conference: Shared Tasks",
    month = nov, year = "2025", address = "Suzhou, China",
    publisher = "Association for Computational Linguistics",
    url = "https://aclanthology.org/2025.arabicnlp-sharedtasks.105/",
    doi = "10.18653/v1/2025.arabicnlp-sharedtasks.105", pages = "762--766"
}

@inproceedings{park2019specaugment,
  title     = {{SpecAugment: A Simple Data Augmentation Method for Automatic Speech Recognition}},
  author    = {Daniel S. Park and William Chan and Yu Zhang and Chung-Cheng Chiu and Barret Zoph and Ekin D. Cubuk and Quoc V. Le},
  year      = {2019},
  booktitle = {{Interspeech 2019}},
  pages     = {2613--2617},
  doi       = {10.21437/Interspeech.2019-2680},
  issn      = {2958-1796},
}

@inproceedings{abdul2021arbert,
    title = "{ARBERT} {\&} {MARBERT}: Deep Bidirectional Transformers for {A}rabic",
    author = "Abdul-Mageed, Muhammad and Elmadany, AbdelRahim and Nagoudi, El Moatez Billah",
    booktitle = "Proceedings of the 59th Annual Meeting of the Association for Computational Linguistics and the 11th International Joint Conference on Natural Language Processing (Volume 1: Long Papers)",
    month = aug, year = "2021", address = "Online",
    publisher = "Association for Computational Linguistics",
    url = "https://aclanthology.org/2021.acl-long.551/",
    doi = "10.18653/v1/2021.acl-long.551", pages = "7088--7105"
}

@inproceedings{elleuch2026ara,
  title={{Ara-BEST-RQ}: Multi Dialectal {Arabic} {SSL}},
  author={Elleuch, Haroun and Whetten, Ryan and Mdhaffar, Salima and Est{\`e}ve, Yannick and Bougares, Fethi},
  booktitle={ICASSP 2026-2026 IEEE International Conference on Acoustics, Speech and Signal Processing (ICASSP)},
  pages={5326--5330},
  year={2026},
  organization={IEEE}
}

@article{dahou2025survey,
  title={A survey on dialect {Arabic} processing and analysis: Recent advances and future trends},
  author={Dahou, Abdelghani and Dahou, Abdelhalim Hafedh and Ch{\'e}ragui, Mohamed Amine and Abdedaiem, Amin and Al-Qaness, Mohammed AA and Abd Elaziz, Mohamed and Ewees, Ahmed A and Zheng, Zhonglong},
  journal={ACM Transactions on Asian and Low-Resource Language Information Processing},
  volume={24},
  number={8},
  pages={1--45},
  year={2025},
  publisher={ACM New York, NY}
}

@article{besdouri2024arabic,
  title={Arabic automatic speech recognition: challenges and progress},
  author={Besdouri, Fatma Zahra and Zribi, In{\`e}s and Belguith, Lamia Hadrich},
  journal={Speech Communication},
  volume={163},
  pages={103110},
  year={2024},
  publisher={Elsevier}
}

@misc{shaun_cassini_2026,
    author       = { Shaun Cassini and Sebastian Vincent and Xiaolu Lu and Julian Mack and Dhruti Joshi and Pierre Richemond },
    title        = { cohere-transcribe-arabic-07-2026 (Revision 0a8193c) },
    year         = 2026,
    url          = { https://huggingface.co/CohereLabs/cohere-transcribe-arabic-07-2026 },
    doi          = { 10.57967/hf/9549 },
    publisher    = { Hugging Face }
}

@book{tur2011spoken,
  title={Spoken language understanding: Systems for extracting semantic information from speech},
  author={Tur, Gokhan and De Mori, Renato},
  year={2011},
  publisher={John Wiley \& Sons}
}

@inproceedings{laperriere2022spoken,
    title = "The Spoken Language Understanding {MEDIA} Benchmark Dataset in the Era of Deep Learning: data updates, training and evaluation tools",
    author = {Laperri{\`e}re, Ga{\"e}lle and Pelloin, Valentin and Caubri{\`e}re, Antoine and Mdhaffar, Salima and Camelin, Nathalie and Ghannay, Sahar and Jabaian, Bassam and Est{\`e}ve, Yannick},
    booktitle = "Proceedings of the Thirteenth Language Resources and Evaluation Conference",
    month = jun, year = "2022", address = "Marseille, France",
    publisher = "European Language Resources Association",
    url = "https://aclanthology.org/2022.lrec-1.171/", pages = "1595--1602"
}

@inproceedings{slurptn,
    title = "{SLURP}-{TN} : Resource for {T}unisian Dialect Spoken Language Understanding",
    author = "Elleuch, Haroun  and
      Mdhaffar, Salima  and
      Est{\`e}ve, Yannick  and
      Bougares, Fethi",
    editor = "Piperidis, Stelios  and
      Bel, N{\'u}ria  and
      van den Heuvel, Henk  and
      Ide, Nancy  and
      Krek, Simon  and
      Toral, Antonio",
    booktitle = "Proceedings of the Fifteenth Language Resources and Evaluation Conference",
    month = may,
    year = "2026",
    address = "Palma de Mallorca, Spain",
    publisher = "ELRA Language Resource Association",
    url = "https://aclanthology.org/2026.lrec-1.119/",
    doi = "10.63317/4m2ac973aco4",
    pages = "1544--1552"
}

@inproceedings{mdhaffar2024taric,
    title = "{TARIC}-{SLU}: A {T}unisian Benchmark Dataset for Spoken Language Understanding",
    author = {Mdhaffar, Salima and Bougares, Fethi and de Mori, Renato and Zaiem, Salah and Ravanelli, Mirco and Est{\`e}ve, Yannick},
    booktitle = "Proceedings of the 2024 Joint International Conference on Computational Linguistics, Language Resources and Evaluation (LREC-COLING 2024)",
    month = may, year = "2024", address = "Torino, Italia",
    publisher = "ELRA and ICCL",
    url = "https://aclanthology.org/2024.lrec-main.1357/", pages = "15606--15616"
}

@inproceedings{mdhaffar2024performance,
    title = "Performance Analysis of Speech Encoders for Low-Resource {SLU} and {ASR} in {T}unisian Dialect",
    author = {Mdhaffar, Salima and Elleuch, Haroun and Bougares, Fethi and Est{\`e}ve, Yannick},
    booktitle = "Proceedings of the Second Arabic Natural Language Processing Conference",
    month = aug, year = "2024", address = "Bangkok, Thailand",
    publisher = "Association for Computational Linguistics",
    url = "https://aclanthology.org/2024.arabicnlp-1.12/",
    doi = "10.18653/v1/2024.arabicnlp-1.12", pages = "130--139"
}

@inproceedings{mdhaffar2025sense,
  title={SENSE models: an open source solution for multilingual and multimodal semantic-based tasks},
  author={Mdhaffar, Salima and Elleuch, Haroun and Chellaf, Chaimae and Nguyen, Ha and Est{\`e}ve, Yannick},
  booktitle={2025 IEEE Automatic Speech Recognition and Understanding Workshop (ASRU)},
  pages={1--8},
  year={2025},
  organization={IEEE}
}

@inproceedings{baevski2023efficient,
  title={Efficient self-supervised learning with contextualized target representations for vision, speech and language},
  author={Baevski, Alexei and Babu, Arun and Hsu, Wei-Ning and Auli, Michael},
  booktitle={International conference on machine learning},
  pages={1416--1429},
  year={2023},
  organization={PMLR}
}

@article{chen2022wavlm,
  title={{WavLM}: Large-scale self-supervised pre-training for full stack speech processing},
  author={Chen, Sanyuan and Wang, Chengyi and Chen, Zhengyang and Wu, Yu and Liu, Shujie and Chen, Zhuo and Li, Jinyu and Kanda, Naoyuki and Yoshioka, Takuya and Xiao, Xiong and others},
  journal={IEEE Journal of Selected Topics in Signal Processing},
  volume={16},
  number={6},
  pages={1505--1518},
  year={2022},
  publisher={IEEE}
}

@inproceedings{lichouri2019arabic,
  title={An {Arabic} multi-domain spoken language understanding system},
  author={Lichouri, Mohamed and Abbas, Mourad and Djeradi, Rachida and Djeradi, Amar},
  booktitle={Proceedings of the 3rd International Conference on Natural Language and Speech Processing},
  pages={49--53},
  year={2019}
}

@inproceedings{lhioui2013combined,
  title={A combined method based on stochastic and linguistic paradigm for the understanding of {Arabic} spontaneous utterances},
  author={Lhioui, Chahira and Zouaghi, Anis and Zrigui, Mounir},
  booktitle={International Conference on Intelligent Text Processing and Computational Linguistics},
  pages={549--558},
  year={2013},
  organization={Springer}
}

@inproceedings{lee24i_interspeech,
  title     = {{Speech-MASSIVE: A Multilingual Speech Dataset for SLU and Beyond}},
  author    = {Beomseok Lee and Ioan Calapodescu and Marco Gaido and Matteo Negri and Laurent Besacier},
  year      = {2024},
  booktitle = {{Interspeech 2024}},
  pages     = {817--821},
  doi       = {10.21437/Interspeech.2024-957},
  issn      = {2958-1796},
}

@inproceedings{fitzgerald2023massive,
    title = "{MASSIVE}: A 1{M}-Example Multilingual Natural Language Understanding Dataset with 51 Typologically-Diverse Languages",
    author = "FitzGerald, Jack and Hench, Christopher and Peris, Charith and Mackie, Scott and Rottmann, Kay and Sanchez, Ana and Nash, Aaron and Urbach, Liam and Kakarala, Vishesh and Singh, Richa and Ranganath, Swetha and Crist, Laurie and Britan, Misha and Leeuwis, Wouter and Tur, Gokhan and Natarajan, Prem",
    booktitle = "Proceedings of the 61st Annual Meeting of the Association for Computational Linguistics (Volume 1: Long Papers)",
    month = jul, year = "2023", address = "Toronto, Canada",
    publisher = "Association for Computational Linguistics",
    url = "https://aclanthology.org/2023.acl-long.235/",
    doi = "10.18653/v1/2023.acl-long.235", pages = "4277--4302"
}

@article{barrault2023seamlessm4t,
  title={{SeamlessM4T}: Massively multilingual \& multimodal machine translation},
  author={Barrault, Lo{\"\i}c and Chung, Yu-An and Meglioli, Mariano Cora and Dale, David and Dong, Ning and Duquenne, Paul-Ambroise and Elsahar, Hady and Gong, Hongyu and Heffernan, Kevin and Hoffman, John and others},
  journal={arXiv preprint arXiv:2308.11596},
  year={2023},
  url = {https://arxiv.org/abs/2308.11596},
}

@inproceedings{radford2023robust,
  title={Robust speech recognition via large-scale weak supervision},
  author={Radford, Alec and Kim, Jong Wook and Xu, Tao and Brockman, Greg and McLeavey, Christine and Sutskever, Ilya},
  booktitle={International conference on machine learning},
  pages={28492--28518},
  year={2023},
  organization={PMLR}
}

@inproceedings{bougares2026whitehouse,
    title = "{W}hite{H}ouse: Translation of the {C}asablanca Corpus for Multi-dialectal {A}rabic Speech Translation",
    author = {Bougares, Fethi and Mdhaffar, Salima and Est{\`e}ve, Yannick},
    booktitle = "Proceedings of the Fifteenth Language Resources and Evaluation Conference",
    month = may, year = "2026", address = "Palma de Mallorca, Spain",
    publisher = "ELRA Language Resource Association",
    url = "https://aclanthology.org/2026.lrec-1.463/",
    doi = "10.63317/4zqn965acien", pages = "5849--5855"
}

@inproceedings{agostinelli-etal-2025-findings,
    title = "Findings of the {IWSLT} 2025 Evaluation Campaign",
    author = {Abdulmumin, Idris  and
      Agostinelli, Victor  and
      Alum{\"a}e, Tanel  and
      Anastasopoulos, Antonios  and
      Bentivogli, Luisa  and
      Bojar, Ond{\v{r}}ej  and
      Borg, Claudia  and
      Bougares, Fethi  and
      Cattoni, Roldano  and
      Cettolo, Mauro  and
      Chen, Lizhong  and
      Chen, William  and
      Dabre, Raj  and
      Est{\`e}ve, Yannick  and
      Federico, Marcello  and
      Fishel, Mark  and
      Gaido, Marco  and
      Javorsk{\'y}, D{\'a}vid  and
      Kasztelnik, Marek  and
      Kponou, Fortun{\'e}  and
      Krubi{\'n}ski, Mateusz  and
      Kin Lam, Tsz  and
      Liu, Danni  and
      Matusov, Evgeny  and
      Kumar Maurya, Chandresh  and
      McCrae, John P.  and
      Mdhaffar, Salima  and
      Moslem, Yasmin  and
      Murray, Kenton  and
      Nakamura, Satoshi  and
      Negri, Matteo  and
      Niehues, Jan  and
      Kr. Ojha, Atul  and
      Ortega, John E.  and
      Papi, Sara  and
      Pecina, Pavel  and
      Pol{\'a}k, Peter  and
      Po{\l}e{\'c}, Piotr  and
      Sankar, Ashwin  and
      Savoldi, Beatrice  and
      Sethiya, Nivedita  and
      Sikasote, Claytone  and
      Sperber, Matthias  and
      St{\"u}ker, Sebastian  and
      Sudoh, Katsuhito  and
      Thompson, Brian  and
      Turchi, Marco  and
      Waibel, Alex  and
      Wilken, Patrick  and
      Zevallos, Rodolfo  and
      Zouhar, Vil{\'e}m  and
      Z{\"u}fle, Maike},
    editor = "Salesky, Elizabeth  and
      Federico, Marcello  and
      Anastasopoulos, Antonis",
    booktitle = "Proceedings of the 22nd International Conference on Spoken Language Translation (IWSLT 2025)",
    month = jul,
    year = "2025",
    address = "Vienna, Austria (in-person and online)",
    publisher = "Association for Computational Linguistics",
    url = "https://aclanthology.org/2025.iwslt-1.44/",
    doi = "10.18653/v1/2025.iwslt-1.44",
    pages = "412--481",
    ISBN = "979-8-89176-272-5"
}

@inproceedings{chellaf2025lia,
    title = "{LIA} and {ELYADATA} systems for the {IWSLT} 2025 low-resource speech translation shared task",
    author = {Chellaf, Chaimae and Elleuch, Haroun and Istaiteh, Othman and Kponou, D. Fortun{\'e} and Bougares, Fethi and Est{\`e}ve, Yannick and Mdhaffar, Salima},
    booktitle = "Proceedings of the 22nd International Conference on Spoken Language Translation (IWSLT 2025)",
    month = jul, year = "2025", address = "Vienna, Austria (in-person and online)",
    publisher = "Association for Computational Linguistics",
    url = "https://aclanthology.org/2025.iwslt-1.27/",
    doi = "10.18653/v1/2025.iwslt-1.27", pages = "274--281"
}

@article{ali2021connecting,
  title={Connecting {Arabs}: Bridging the gap in dialectal speech recognition},
  author={Ali, Ahmed and Chowdhury, Shammur and Afify, Mohamed and El-Hajj, Wassim and Hajj, Hazem and Abbas, Mourad and Hussein, Amir and Ghneim, Nada and Abushariah, Mohammad and Alqudah, Assal},
  journal={Communications of the ACM},
  volume={64},
  number={4},
  pages={124--129},
  year={2021},
  publisher={ACM New York, NY, USA}
}

@inproceedings{ali2016mgb,
  title={The {MGB-2} challenge: {Arabic} multi-dialect broadcast media recognition},
  author={Ali, Ahmed and Bell, Peter and Glass, James and Messaoui, Yacine and Mubarak, Hamdy and Renals, Steve and Zhang, Yifan},
  booktitle={2016 IEEE Spoken Language Technology Workshop (SLT)},
  pages={279--284},
  year={2016},
  organization={IEEE}
}

@inproceedings{mubarak-etal-2021-qasr,
    title = "{QASR}: {QCRI} {Aljazeera} Speech Resource A Large Scale Annotated {A}rabic Speech Corpus",
    author = "Mubarak, Hamdy  and
      Hussein, Amir  and
      Chowdhury, Shammur Absar  and
      Ali, Ahmed",
    editor = "Zong, Chengqing  and
      Xia, Fei  and
      Li, Wenjie  and
      Navigli, Roberto",
    booktitle = "Proceedings of the 59th Annual Meeting of the Association for Computational Linguistics and the 11th International Joint Conference on Natural Language Processing (Volume 1: Long Papers)",
    month = aug,
    year = "2021",
    address = "Online",
    publisher = "Association for Computational Linguistics",
    url = "https://aclanthology.org/2021.acl-long.177/",
    doi = "10.18653/v1/2021.acl-long.177",
    pages = "2274--2285"
}

@inproceedings{ali2015multi,
  title={Multi-reference {WER} for evaluating {ASR} for languages with no orthographic rules},
  author={Ali, Ahmed and Magdy, Walid and Bell, Peter and Renais, Steve},
  booktitle={2015 IEEE Workshop on Automatic Speech Recognition and Understanding (ASRU)},
  pages={576--580},
  year={2015},
  organization={IEEE}
}

@inproceedings{laurent2023trac,
    title = "{ON}-{TRAC} Consortium Systems for the {IWSLT} 2023 Dialectal and Low-resource Speech Translation Tasks",
    author = {Laurent, Antoine and Gahbiche, Souhir and Nguyen, Ha and Elleuch, Haroun and Bougares, Fethi and Thiol, Antoine and Riguidel, Hugo and Mdhaffar, Salima and Laperri{\`e}re, Ga{\"e}lle and Maison, Lucas and Khurana, Sameer and Est{\`e}ve, Yannick},
    booktitle = "Proceedings of the 20th International Conference on Spoken Language Translation (IWSLT 2023)",
    month = jul, year = "2023", address = "Toronto, Canada (in-person and online)",
    publisher = "Association for Computational Linguistics",
    url = "https://aclanthology.org/2023.iwslt-1.18/",
    doi = "10.18653/v1/2023.iwslt-1.18", pages = "219--226"
}

@inproceedings{bougares2025tedxtn,
    title = "{TED}x{TN}: A Three-way Speech Translation Corpus for Code-Switched {T}unisian {A}rabic - {E}nglish",
    author = {Bougares, Fethi and Mdhaffar, Salima and Elleuch, Haroun and Est{\`e}ve, Yannick},
    booktitle = "Proceedings of The Third Arabic Natural Language Processing Conference",
    month = nov, year = "2025", address = "Suzhou, China",
    publisher = "Association for Computational Linguistics",
    url = "https://aclanthology.org/2025.arabicnlp-main.22/",
    doi = "10.18653/v1/2025.arabicnlp-main.22", pages = "278--287"
}

@inproceedings{hamed2022arzen,
    title = "{A}rz{E}n-{ST}: A Three-way Speech Translation Corpus for Code-Switched {E}gyptian {A}rabic-{E}nglish",
    author = "Hamed, Injy and Habash, Nizar and Abdennadher, Slim and Vu, Ngoc Thang",
    booktitle = "Proceedings of the Seventh Arabic Natural Language Processing Workshop (WANLP)",
    month = dec, year = "2022", address = "Abu Dhabi, United Arab Emirates (Hybrid)",
    publisher = "Association for Computational Linguistics",
    url = "https://aclanthology.org/2022.wanlp-1.12/",
    doi = "10.18653/v1/2022.wanlp-1.12", pages = "119--130"
}

@inproceedings{hamed2020arzen,
    title = "{A}rz{E}n: A Speech Corpus for Code-switched {E}gyptian {A}rabic-{E}nglish",
    author = "Hamed, Injy and Vu, Ngoc Thang and Abdennadher, Slim",
    booktitle = "Proceedings of the Twelfth Language Resources and Evaluation Conference",
    month = may, year = "2020", address = "Marseille, France",
    publisher = "European Language Resources Association",
    url = "https://aclanthology.org/2020.lrec-1.523/",
    pages = "4237--4246", isbn = "979-10-95546-34-4"
}

@inproceedings{sabaa-nadi2026,
    title = "{Sabaa at {NADI} 2026 Subtask 1.1: Matching Annotation Conventions for Dialectal {Arabic} {ASR}}",
    author = {Mohamed Sabaa and Amr Sabaa},
    booktitle = "Proceedings of the Fourth Arabic Natural Language Processing Conference (ArabicNLP 2026)",
    year = "2026",
    address = "Budapest, Hungary",
    publisher = "Association for Computational Linguistics",
}

@inproceedings{aslema-nadi2026,
    title = "{Aslema at {NADI} 2026: Data Augmentation for Intent Recognition and Slot Filling}",
    author = {Tajwaar Shafiq and Hunzalah Hassan Bhatti and  Firoj Alam and Shammur Absar Chowdhury},
    booktitle = "Proceedings of the Fourth Arabic Natural Language Processing Conference (ArabicNLP 2026)",
    year = "2026",
    address = "Budapest, Hungary",
    publisher = "Association for Computational Linguistics",
}

@inproceedings{abjad-nadi2026,
    title = "{Abjad {AI} at {NADI} 2026 shared task: Overcoming Data Scarcity in {Arabic} Speech Processing through Augmentation}",
    author = {
    Hatim Alhumid and Naif Alharthi and Elaf S. Alsaedi and Faris Alasmary and Ahmad Ghannam and Anas Adel Salamah and Shouq Sadah and Lahouari Ghouti
    },
    booktitle = "Proceedings of the Fourth Arabic Natural Language Processing Conference (ArabicNLP 2026)",
    year = "2026",
    address = "Budapest, Hungary",
    publisher = "Association for Computational Linguistics",
}

@inproceedings{nile-nadi2026,
    title = "{A Two-Phase Acoustic Fine-Tuning Strategy for Robust Dialectal {Arabic} Speech Recognition {NADI} 2026 Subtask 1.1: Noisy Country-Level {ASR}}",
    author = {Mahmoud  Alshrief and Mohamed Bahgat and Mahmoud Zaher and Sahar Selim},
    booktitle = "Proceedings of the Fourth Arabic Natural Language Processing Conference (ArabicNLP 2026)",
    year = "2026",
    address = "Budapest, Hungary",
    publisher = "Association for Computational Linguistics",
}

@inproceedings{wifaq-nadi2026,
    title = "{Wifaq at {NADI} 2026 Shared Task: Diversity-Aware System Combination for Robust {Arabic} {ASR}}",
    author = {Sulaiman Alkurbi},
    booktitle = "Proceedings of the Fourth Arabic Natural Language Processing Conference (ArabicNLP 2026)",
    year = "2026",
    address = "Budapest, Hungary",
    publisher = "Association for Computational Linguistics",
}

@inproceedings{namaa-nadi2026,
    title = "{{NAMAA Community} at {NADI} 2026: Mixed-Dialect and Code-Switched {Arabic} {ASR} and Spoken Dialect Identification}",
    author = {Mohamed Abdelazi and Fatimah Mohamed Emad Elden and Mousa Ashraf Abd El Malak and Omer Nacar},
    booktitle = "Proceedings of the Fourth Arabic Natural Language Processing Conference (ArabicNLP 2026)",
    year = "2026",
    address = "Budapest, Hungary",
    publisher = "Association for Computational Linguistics",
}

@inproceedings{cis_openslu-nadi2026,
    title = "{{CIS-OpenSLU} at {NADI} 2026 Shared Task: {ASR}-Matched Training and Confidence-Based Rejection for Open-Set {Tunisian} {Arabic} Intent Recognition}",
    author = {Salma Khaled},
    booktitle = "Proceedings of the Fourth Arabic Natural Language Processing Conference (ArabicNLP 2026)",
    year = "2026",
    address = "Budapest, Hungary",
    publisher = "Association for Computational Linguistics",
}

@inproceedings{codezone-nadi2026,
    title = "{{CodeZone} Research Group at {NADI} 2026 Shared Task: Cross-Domain Score Calibration for Spoken {Arabic} Dialect Identification}",
    author = {Sarah Yassine and Abdulkadir Shehu Bichi},
    booktitle = "Proceedings of the Fourth Arabic Natural Language Processing Conference (ArabicNLP 2026)",
    year = "2026",
    address = "Budapest, Hungary",
    publisher = "Association for Computational Linguistics",
}

@inproceedings{ai_elites-nadi2026,
    title = "{{AI Elites} at {NADI} 2026 Shared Task: Selecting on Out-of-Domain Accuracy for Cross-Domain Spoken {Arabic} Dialect Identification}",
    author = {Musab Iskandar and Mohamed Ebrahim and  Adel Alharbi and Khaled Alharbi and Youssef Mohamed},
    booktitle = "Proceedings of the Fourth Arabic Natural Language Processing Conference (ArabicNLP 2026)",
    year = "2026",
    address = "Budapest, Hungary",
    publisher = "Association for Computational Linguistics",
}

@inproceedings{thakaa-nadi2026,
    title = "{Thakaa at {NADI} 2026 Shared Task: {ROVER} Ensembling and Multi-Representation Fusion for Robust {Arabic} {ASR} and Dialect Identification}",
    author = {Abdulaziz Otaif and Hazem Bakhshwain and Abdulhamid Aldoobi and Meshal Alamr and Abdullah Aldahlawi},
    booktitle = "Proceedings of the Fourth Arabic Natural Language Processing Conference (ArabicNLP 2026)",
    year = "2026",
    address = "Budapest, Hungary",
    publisher = "Association for Computational Linguistics",
}

@inproceedings{kand-nadi2026,
    title = "{{KAND CA} at {NADI} 2026 Shared Task: Multi-Stage Fine-Tuning for Spoken {Arabic} Dialects}",
    author = {Ahmed Wasfy and Nourhan Anber},
    booktitle = "Proceedings of the Fourth Arabic Natural Language Processing Conference (ArabicNLP 2026)",
    year = "2026",
    address = "Budapest, Hungary",
    publisher = "Association for Computational Linguistics",
}

@inproceedings{lynx-nadi2026,
    title = "{Lynx at the {NADI} 2026 Shared Task: Exploring {ASR} Representations and Transcripts for Dialectal {Arabic} Speech Understanding}",
    author = {Houdaifa Atou and Issam Ait Yahia and Ismail Berrada},
    booktitle = "Proceedings of the Fourth Arabic Natural Language Processing Conference (ArabicNLP 2026)",
    year = "2026",
    address = "Budapest, Hungary",
    publisher = "Association for Computational Linguistics",
}

@inproceedings{resonate-nadi2026,
    title = "{Resonate at {NADI} 2026 Shared Task: Robust Speech Recognition and Dialect Identification under Acoustic and Domain Shift}",
    author = {Omar Shoaib and Mohamed Sokar and Mohamed Motawie and  Omar AboElFottouh and Ahmed Hassouna and AlAmir Hassan and Nader Essam and Wael Ali},
    booktitle = "Proceedings of the Fourth Arabic Natural Language Processing Conference (ArabicNLP 2026)",
    year = "2026",
    address = "Budapest, Hungary",
    publisher = "Association for Computational Linguistics",
}

@inproceedings{itgan-nadi2026,
    title = "{Itgan at {NADI} 2026 shared task: Parameter-Efficient {Whisper} Adaptation for Robust, Mixed-Dialect and Code-Switched {Arabic} {ASR}}",
    author = {Ibrahim Almajai},
    booktitle = "Proceedings of the Fourth Arabic Natural Language Processing Conference (ArabicNLP 2026)",
    year = "2026",
    address = "Budapest, Hungary",
    publisher = "Association for Computational Linguistics",
}

@inproceedings{sacrebleu,
    title = "A Call for Clarity in Reporting {BLEU} Scores",
    author = "Post, Matt",
    editor = "Bojar, Ond{\v{r}}ej  and
      Chatterjee, Rajen  and
      Federmann, Christian  and
      Fishel, Mark  and
      Graham, Yvette  and
      Haddow, Barry  and
      Huck, Matthias  and
      Yepes, Antonio Jimeno  and
      Koehn, Philipp  and
      Monz, Christof  and
      Negri, Matteo  and
      N{\'e}v{\'e}ol, Aur{\'e}lie  and
      Neves, Mariana  and
      Post, Matt  and
      Specia, Lucia  and
      Turchi, Marco  and
      Verspoor, Karin",
    booktitle = "Proceedings of the Third Conference on Machine Translation: Research Papers",
    month = oct,
    year = "2018",
    address = "Brussels, Belgium",
    publisher = "Association for Computational Linguistics",
    url = "https://aclanthology.org/W18-6319/",
    doi = "10.18653/v1/W18-6319",
    pages = "186--191"
}

@inproceedings{laouirine2024tunartts,
    title = "{T}un{A}r{TTS}: {T}unisian {A}rabic Text-To-Speech Corpus",
    author = "Laouirine, Imen and Kammoun, Rami and Bougares, Fethi",
    booktitle = "Proceedings of the 2024 Joint International Conference on Computational Linguistics, Language Resources and Evaluation (LREC-COLING 2024)",
    month = may, year = "2024", address = "Torino, Italia",
    publisher = "ELRA and ICCL",
    url = "https://aclanthology.org/2024.lrec-main.1467/", pages = "16879--16889"
}

@inproceedings{bouchakour2025enhancing,
  title={Enhancing {Arabic} {ASR} in Noisy and Transcoding {EVS} Conditions: A Multimodal Deep Learning Study.},
  author={Bouchakour, Lallouani and Lounnas, Khaled and Krobba, Ahmed},
  booktitle={FedCSIS (Position Papers)},
  pages={1--8},
  year={2025}
}

@inproceedings{cite_bulbul,
title = "Bulbul: A Dataset for Dialectal {Arabic} Speech Recognition",
author = "Ashraf, Ahmed and
Alansari, Aisha and
Al Abbas, Fadel and
Almarwani, Nada and
Aloufi, Samah and
Ezzini, Saad and
Al-shaibani, Maged S. and
Dalaq, Doaa and
Elmadany, AbdelRahim A. and
Abdul-Mageed, Muhammad and
Trigui, Mohamed Mehdi and
Refai, Dania and
Refai, Layan and
Akrout, Mohamed and
Jarrar, Mustafa and
Al-Khatib, Wasfi G. and
Dalaq, Alaa and
El-Nakla, Darin and
Abdaljalil, Samir and
Al-Fakih, Abdulrahman and
Zeghib, Nour El Imane and
Redah, Moussa and
Chafik, Salmane and
El-Attar, Mohamed and
Grati, Rima and
Kohail, Sarah and
Alkhorasani, Malak and
Al Safwan, Khadijah and
Mudhaffar, Ismail Mohmmed and
Altam, Ali and
Al-Shaikh, Ahmed and
Saeed, Adnan and
Luqman, Hamzah",
booktitle = "Proceedings of the 2026 Conference on Empirical Methods in Natural Language Processing",
month = oct,
year = "2026",
address = "Budapest, Hungary",
publisher = "Association for Computational Linguistics",
pages = "",
note = "To appear",
}

@inproceedings{sullivan-etal-2026-arab,
    title = "{A}rab Voices: Mapping Standard and Dialectal {A}rabic Speech Technology",
    author = "Sullivan, Peter  and
      Elmadany, AbdelRahim A.  and
      Alcoba Inciarte, Alcides  and
      Abdul-Mageed, Muhammad",
    editor = "Liakata, Maria  and
      Moreira, Viviane P.  and
      Zhang, Jiajun  and
      Jurgens, David",
    booktitle = "Findings of the {A}ssociation for {C}omputational {L}inguistics: {ACL} 2026",
    month = jul,
    year = "2026",
    address = "San Diego, California, United States",
    publisher = "Association for Computational Linguistics",
    url = "https://aclanthology.org/2026.findings-acl.575/",
    doi = "10.18653/v1/2026.findings-acl.575",
    pages = "11843--11878",
    ISBN = "979-8-89176-395-1"
}

@article{yang2025qwen3,
  title={Qwen3 technical report},
  author={Yang, An and Li, Anfeng and Yang, Baosong and Zhang, Beichen and Hui, Binyuan and Zheng, Bo and Yu, Bowen and Gao, Chang and Huang, Chengen and Lv, Chenxu and others},
  journal={arXiv preprint arXiv:2505.09388},
  year={2025},
  url = {https://arxiv.org/abs/2505.09388},
}

@inproceedings{malmasi2016arabic,
    title = "{A}rabic Dialect Identification in Speech Transcripts",
    author = "Malmasi, Shervin and Zampieri, Marcos",
    booktitle = "Proceedings of the Third Workshop on {NLP} for Similar Languages, Varieties and Dialects ({V}ar{D}ial3)",
    month = dec, year = "2016", address = "Osaka, Japan",
    publisher = "The COLING 2016 Organizing Committee",
    url = "https://aclanthology.org/W16-4814/", pages = "106--113"
}

@inproceedings{biadsy2009spoken,
  title={Spoken {Arabic} dialect identification using phonotactic modeling},
  author={Biadsy, Fadi and Hirschberg, Julia and Habash, Nizar},
  booktitle={Proceedings of the EACL 2009 workshop on computational approaches to semitic languages},
  pages={53--61},
  year={2009},
  url = {https://aclanthology.org/W09-0807/},
}

@inproceedings{klejch2025practitioner,
  title     = {{A Practitioner’s Guide to Building ASR Models for Low-Resource Languages: A Case Study on Scottish Gaelic}},
  author    = {Ondřej Klejch and William Lamb and Peter Bell},
  year      = {2025},
  booktitle = {{Interspeech 2025}},
  pages     = {728--732},
  doi       = {10.21437/Interspeech.2025-1557},
  issn      = {2958-1796},
}

@inproceedings{choux2025tunifra,
    title = "{T}uni{F}ra: A {T}unisian {A}rabic Speech Corpus with Orthographic Transcriptions and {F}rench Translations",
    author = "Choux, Alex and Avila, Marko and Crego, Josep and Bougares, Fethi and Laurent, Antoine",
    booktitle = "Proceedings of The Third Arabic Natural Language Processing Conference",
    month = nov, year = "2025", address = "Suzhou, China",
    publisher = "Association for Computational Linguistics",
    url = "https://aclanthology.org/2025.arabicnlp-main.5/",
    doi = "10.18653/v1/2025.arabicnlp-main.5", pages = "64--68"
}

@inproceedings{fiscus1997post,
  title={A post-processing system to yield reduced word error rates: Recognizer output voting error reduction ({ROVER})},
  author={Fiscus, Jonathan G},
  booktitle={1997 IEEE workshop on automatic speech recognition and understanding proceedings},
  pages={347--354},
  year={1997},
  organization={IEEE}
}

@inproceedings{heafield2011kenlm,
    title = "{K}en{LM}: Faster and Smaller Language Model Queries",
    author = "Heafield, Kenneth",
    booktitle = "Proceedings of the Sixth Workshop on Statistical Machine Translation",
    month = jul, year = "2011", address = "Edinburgh, Scotland",
    publisher = "Association for Computational Linguistics",
    url = "https://aclanthology.org/W11-2123/", pages = "187--197"
}

@inproceedings{xue2021mt5,
    title = "m{T5}: A Massively Multilingual Pre-trained Text-to-Text Transformer",
    author = "Xue, Linting and Constant, Noah and Roberts, Adam and Kale, Mihir and Al-Rfou, Rami and Siddhant, Aditya and Barua, Aditya and Raffel, Colin",
    booktitle = "Proceedings of the 2021 Conference of the North American Chapter of the Association for Computational Linguistics: Human Language Technologies",
    month = jun, year = "2021", address = "Online",
    publisher = "Association for Computational Linguistics",
    url = "https://aclanthology.org/2021.naacl-main.41/",
    doi = "10.18653/v1/2021.naacl-main.41", pages = "483--498"
}

@article{zhu2026omnivoice,
  title={{OmniVoice}: Towards omnilingual zero-shot text-to-speech with diffusion language models},
  author={Zhu, Han and Ye, Lingxuan and Kang, Wei and Yao, Zengwei and Guo, Liyong and Kuang, Fangjun and Han, Zhifeng and Zhuang, Weiji and Lin, Long and Povey, Daniel},
  journal={arXiv preprint arXiv:2604.00688},
  year={2026},
  url = {https://arxiv.org/abs/2604.00688},
}

@article{omnilingual2025omnilingual,
  title={Omnilingual {ASR}: Open-source multilingual speech recognition for 1600+ languages},
  author={{Omnilingual ASR} and Keren, Gil and Kozhevnikov, Artyom and Meng, Yen and Ropers, Christophe and Setzler, Matthew and Wang, Skyler and Adebara, Ife and Auli, Michael and Balioglu, Can and others},
  journal={arXiv preprint arXiv:2511.09690},
  year={2025},
  url = {https://arxiv.org/abs/2511.09690},
}

@misc{moulsot2026,
  title   = {{MoulSot}: A Curated {Moroccan} {Darija} Speech Dataset and Fine-Tuned {ASR} Model},
  author  = {{Atlasia}},
  year    = {2026},
  url     = {https://huggingface.co/atlasia/moulsot.v0.3}
}

@inproceedings{zhao25f_interspeech,
  title     = {{ClearerVoice-Studio: Bridging Advanced Speech Processing Research and Practical Deployment}},
  author    = {Shengkui Zhao and Zexu Pan and Bin Ma},
  year      = {2025},
  booktitle = {{Interspeech 2025}},
  pages     = {2980--2984},
  doi       = {10.21437/Interspeech.2025-680},
  issn      = {2958-1796},
}

@inproceedings{rekesh2023fast,
  title={Fast {C}onformer with linearly scalable attention for efficient speech recognition},
  author={Rekesh, Dima and Koluguri, Nithin Rao and Kriman, Samuel and Majumdar, Somshubra and Noroozi, Vahid and Huang, He and Hrinchuk, Oleksii and Puvvada, Krishna and Kumar, Ankur and Balam, Jagadeesh and others},
  booktitle={2023 IEEE Automatic Speech Recognition and Understanding Workshop (ASRU)},
  pages={1--8},
  year={2023},
  organization={IEEE}
}

@misc{kandca_lag,
author={{KAND CA}},
title={Lahgtna Huggingface Dataset},
year=2026,
url={https://huggingface.co/datasets/oddadmix/lahgtna-v3-small}
}

@inproceedings{conneau2023fleurs,
  title={{FLEURS}: Few-shot learning evaluation of universal representations of speech},
  author={Conneau, Alexis and Ma, Min and Khanuja, Simran and Zhang, Yu and Axelrod, Vera and Dalmia, Siddharth and Riesa, Jason and Rivera, Clara and Bapna, Ankur},
  booktitle={2022 IEEE Spoken Language Technology Workshop (SLT)},
  pages={798--805},
  year={2023},
  organization={IEEE}
}

@misc{SileroVAD,
  author = {{Silero Team}},
  title = {Silero {VAD}: pre-trained enterprise-grade Voice Activity Detector ({VAD}), Number Detector and Language Classifier},
  year = {2024},
  publisher = {GitHub},
  journal = {GitHub repository},
  howpublished = {\url{https://github.com/snakers4/silero-vad}},
}

@inproceedings{piczak2015esc,
  title={{ESC}: Dataset for environmental sound classification},
  author={Piczak, Karol J},
  booktitle={Proceedings of the 23rd ACM international conference on Multimedia},
  pages={1015--1018},
  year={2015}
}

@article{zhou2026voxcpm2,
  title={{VoxCPM2} technical report},
  author={Zhou, Yixuan and Zeng, Guoyang and Liu, Xin and Li, Xiang and Yu, Renjie and Gui, Jiancheng and Wu, Jiaheng and Wang, Ziyang and Shen, Xudong and Ye, Runchuan and others},
  journal={arXiv preprint arXiv:2606.06928},
  year={2026},
  url = {https://arxiv.org/abs/2606.06928},
}

@inproceedings{popovic-2015-chrf,
    title = "chr{F}: character n-gram {F}-score for automatic {MT} evaluation",
    author = "Popovi{\'c}, Maja",
    editor = "Bojar, Ond{\v{r}}ej  and
      Chatterjee, Rajan  and
      Federmann, Christian  and
      Haddow, Barry  and
      Hokamp, Chris  and
      Huck, Matthias  and
      Logacheva, Varvara  and
      Pecina, Pavel",
    booktitle = "Proceedings of the Tenth Workshop on Statistical Machine Translation",
    month = sep,
    year = "2015",
    address = "Lisbon, Portugal",
    publisher = "Association for Computational Linguistics",
    url = "https://aclanthology.org/W15-3049/",
    doi = "10.18653/v1/W15-3049",
    pages = "392--395"
}

@inproceedings{papineni2002bleu,
  title={{BLEU}: a method for automatic evaluation of machine translation},
  author={Papineni, Kishore and Roukos, Salim and Ward, Todd and Zhu, Wei-Jing},
  booktitle={Proceedings of the 40th annual meeting of the Association for Computational Linguistics},
  pages={311--318},
  year={2002}
}

\clearpage
\appendix
\addappheadtotoc         
\numberwithin{figure}{section}
\numberwithin{table}{section}

\section*{Appendix Table Of Contents}

\begin{itemize}
    \item  Comprehensive Literature Review: \S \ref{sec:appendix_lit}
    \item  Additional Task Information: \S \ref{sec:appendix_task_supp}
    \item  Baseline Systems: \S \ref{sec:appendix_baselines}
    \item  Additional Result Tables: \S \ref{sec:appendix_results}
    \item  Complete Submission Descriptions: \S \ref{sec:appendix_systems}
\end{itemize}

\section*{List of Tables}
\begin{itemize}
    \item  Overview NADI Shared Tasks: Table \ref{tab:nadi_evolution_full}

    \item  Participation Overview: Table \ref{tab:participation}
    \item  TTS Country Dist.: Table \ref{tab:country_samples}
    \item  SLT Data Overview: Table \ref{tab:whitehouse_test}
    \item Subtask 1.3 CS ASR Results: Table \ref{tab:nadi26-subtask13-systems}
    \item Subtask 2 SDID Results: Table \ref{tab:adi-2-res}
    \item Subtask 3 TTS Results: Table \ref{tab:tts-task3-results}
    \item Subtask 4 SLT Results: Table \ref{tab:ast-task4-results}
    \item Subtask 5.1 Intent Class. Results: Table \ref{tab:slu_5_1_res}
    \item Subtask 5.2 Slot Filling Results: Table \ref{tab:slu-5-2-res}
\end{itemize}

\section{Comprehensive Literature Review}
\label{sec:appendix_lit}

\subsection{Automatic Speech Recognition}
\paragraph{Foundations of ASR.}
Automatic speech recognition has progressed from statistical HMM-GMM systems~\cite{rabiner1989tutorial} and hybrid DNN-HMM acoustic models~\cite{hinton2012deep} to end-to-end architectures trained with connectionist temporal classification~\cite{graves2006connectionist} and attention-based encoder-decoders~\cite{chan2016listen}. The Conformer~\cite{gulati2020conformer} combined convolution and self-attention as a standard encoder backbone, while self-supervised pretraining~\cite{baevski2020wav2vec, hsu2021hubert} and large-scale weakly supervised training~\cite{radford2023robust} substantially reduced the field's dependence on labeled data. These advances underpin nearly all recent dialectal and low-resource ASR work, including on Arabic, where Whisper- and MMS-based models~\cite{radford2023robust, pratap2024scaling} now serve as the default starting point for dialect-specific adaptation~\cite{wang2024open, ozyilmaz2025overcoming}.

\paragraph{Low-Resource and Multilingual ASR.}
A substantial line of work targets languages and varieties with limited labeled data. Approaches include transfer learning from phonetically similar high-resource languages combined with community-sourced lexical resources~\cite{kopikar2026amchi}, self-supervised pretraining on large unlabeled corpora followed by minimal fine-tuning~\cite{zhuo2025vietasr}, and hybrid HMM/self-supervised pipelines that outperform na\"{i}vely fine-tuned end-to-end models~\cite{klejch2025practitioner}. In-context learning with LLMs has also been shown to acquire unseen low-resource languages without dedicated ASR training~\cite{li2025context}, and self-supervised encoders have been scaled to over a thousand African languages~\cite{alabi2024afrihubert}. Arabic dialects sit squarely within this low-resource paradigm: despite being spoken by hundreds of millions of people, most dialectal varieties lack the transcribed corpora available for MSA~\cite{ali2017speech, madar-2018-bouamor}, motivating dedicated efforts such as the NADI shared task series~\cite{talafha-etal-2025-nadi}.

\paragraph{Dialectal, Accented, and Sociolinguistic Robustness.}
ASR systems trained predominantly on standard or high-resource varieties degrade substantially on dialectal and accented speech. This has been documented for German dialects~\cite{blaschke2025multi}, Chinese dialects~\cite{xu2025leveraging}, regional English varieties such as Newcastle English~\cite{serditova2025automatic}, and African American English~\cite{mojarad2025automatic}. Proposed mitigations range from geographic-metadata conditioning~\cite{mehralian2025leveraging} to mixtures of accent-specific LoRA experts~\cite{bagat2025mixture}. Arabic exhibits some of the most pronounced dialectal variation of any widely spoken language, and a growing body of work specifically fine-tunes and benchmarks multilingual ASR models across Arabic dialects~\cite{wang2024open, ozyilmaz2025overcoming}, jointly modeling dialect identification alongside recognition~\cite{kumar2026jointly}. Code-switching compounds this variation further, and while most code-switching ASR work has focused on non-Arabic language pairs~\cite{biswas2025adapting, yang2025adapting}, Arabic speech routinely mixes dialect, MSA, and other languages, a setting NADI 2026 targets directly.

\paragraph{Speech-LLM Integration and Error Correction.}
Recent work couples ASR with large language models for contextual biasing~\cite{gong2025br}, retrieval-augmented correction~\cite{wang2025rag}, and multi-ASR ensemble arbitration~\cite{prakash2025better}, extending classic hypothesis-fusion approaches with LLM-based reasoning. These techniques are largely language-agnostic and have primarily been validated on high-resource languages; their application to dialectal Arabic, where named entities, code-switched terms, and non-standard orthography are common, remains comparatively unexplored.


\subsection{Spoken Language and Dialect ID}
Spoken language identification and its cousin spoken dialect identification have largely transitioned from traditional approaches such as i-vector~\cite{dehak2011language} and x-vector~\cite{snyder2018spoken} to end-to-end deep neural net models often using a transformer~\cite{vaswani2017attention,babu22_interspeech} or ECAPA-TDNN~\cite{desplanques20_interspeech} backbone. Arabic dialect identification has similarly followed these trends moving from i-vector approaches during the MGB-3 shared task~\cite{ali2017speech, shon2017qcri}, to end-to-end neural nets trained from scratch in the follow-up MGB-5 shared task~\cite{ali2019mgb}, and finetuning pretrained models for dialect identification in NADI 2025~\cite{elleuch2025elyadata,talafha-etal-2025-nadi}.

One consistent issue with spoken language identification is the problem of cross domain performance~\cite{sullivan23_interspeech,abdullah25_interspeech,kang2022deep,dey2023cross}. Models may overfit to training domains, rendering them ineffective in real-world scenarios.

\subsection{Text-to-Speech.}
Driven by recent breakthroughs in autoregressive, diffusion-based, and zero-shot architectures, modern TTS systems have demonstrated remarkable naturalness and expressiveness~\cite{tan2021surveyneuralspeechsynthesis}. However, while high-resource languages—and models such as Gemini-TTS~\cite{googlecloud_gemini_tts} or Qwen-based speech architectures~\cite{xiang2026qwenaudio30ttsfreelycontrollablehighly}—have advanced significantly, progress in the Arabic language remains constrained. The primary bottleneck lies in orthographic ambiguity (e.g., the omission of diacritics) and the rich morphological and dialectal variations across the Arab world. To provide a structured overview of existing Arabic TTS capabilities, we categorize existing speech synthesis systems into three primary groups: multilingual models, MSA models, and DA models.

    \paragraph{Multilingual Models} Broad-coverage foundational speech models, such as XTTS~\cite{casanova2024xttsmassivelymultilingualzeroshot}, SeamlessM4T~\cite{barrault2023seamlessm4t}, SpeechT5~\cite{ao2022speecht5unifiedmodalencoderdecoderpretraining}, and ElevenLabs Multilingual v2~\cite{elevenlabs_multilingual_v2}, support Arabic within unified cross-lingual representations. However, they struggle with Arabic phonetic subtleties, orthographic ambiguity, and dialect-specific prosody. Furthermore, because these models do not allow explicit conditioning on specific regional dialects, their ability to synthesize authentic local varieties remains limited-highlighting a critical gap in existing multilingual TTS architectures.
    
    \paragraph{MSA Models} Dedicated acoustic and neural synthesis architectures trained specifically on diacritized formal corpora. Representative examples include ArTST~\citep{toyin2023artst}, and SILMA TTS MSA~\cite{silma_arabic_tts}, alongside commercial engines like Amazon Polly (\textit{Zeina})~\cite{smolarek2019amazonpolly}. While these models achieve high intelligibility and naturalness in formal registers and news broadcast domains, they fail to generalize to informal, colloquial speech patterns due to distinct dialectal vocabulary, syntax, and phonology.
    
   \paragraph{DA Models} Emerging architectures and fine-tuned models specifically tailored to capture country- and region-specific phonetic, rhythmic, and prosodic characteristics. Notable examples include NileTTS~\cite{Khamis_2026}, EGTTS~\cite{OmarSamir}, Habibi-TTS~\cite{chen2026habibilayingopensourcefoundation}, and lahgtna-chatterbox-v1~\cite{oddadmix_lahgtna_chatterbox}, which target regional varieties such as Egyptian, Levantine, and Gulf Arabic. Despite their necessity for realistic conversational AI, dialectal TTS systems remain severely under-resourced compared to English and MSA due to the scarcity of high-quality, paired dialectal speech-text corpora.

\subsection{Spoken Language Translation}

\paragraph{Multilingual Models} Large-scale multilingual speech translation models provide the most developed approach for Arabic SLT. 
Representative systems include SeamlessM4T~\cite{barrault2023seamlessm4t} and Whisper~\cite{radford2023robust}, which benefit from multilingual speech pretraining and can support Arabic as either a source or target language.
More recent foundation models such as SeamlessM4T v2~\cite{barrault2023seamless} further improve multilingual speech-to-text and speech-to-speech translation capabilities. However, these models are primarily optimized for broad multilingual coverage and may struggle to preserve dialect-specific lexical, phonological, and prosodic characteristics, especially for low-resource Arabic varieties.



\paragraph{DA Models} Spoken translation for dialectal Arabic remains considerably under-resourced compared with MSA and high-resource languages. 
Recent efforts have increasingly focused on developing speech--translation resources for Arabic dialects. 
In particular, the WhiteHouse corpus~\cite{bougares2026whitehouse} was recently introduced as the first multi-dialectal Arabic--English speech translation corpus, providing three-way parallel speech, transcription, and English translation data across multiple Arabic dialects. For Tunisian Arabic, the IWSLT Dialectal Speech Translation shared task has provided dedicated resources for Tunisian--English speech translation, including 160 hours~\cite{agostinelli-etal-2025-findings}. 
These resources have enabled the development and evaluation of dialect-specific speech translation systems~\cite{chellaf2025lia,laurent2023trac}, while challenges related to dialectal variation, code-switching, spontaneous speech, and limited parallel supervision remain largely open.
In addition, TEDxTN~\cite{bougares2025tedxtn} is a three-way code-switched Tunisian Arabic--English speech translation corpus comprising 108 TEDx talks and approximately 25 hours of speech from speakers representing more than 11 regions of Tunisia. The corpus provides manually validated transcriptions and English translations at the utterance level, making it a valuable resource for studying code-switched and regionally diverse Tunisian speech translation.

For Egyptian Arabic, ArzEnST~\cite{hamed2022arzen} is a three-way Egyptian Arabic--English code-switched speech translation corpus derived from the ArzEn speech corpus~\cite{hamed2020arzen}. It consists of approximately 12 hours of speech and 6,216 sentences collected through informal interviews with bilingual speakers. The corpus provides speech, Egyptian Arabic transcriptions, and English translations, enabling the development of cascaded speech translation systems in which an ASR model first transcribes the speech and an MT model subsequently translates the resulting transcripts.

\subsection{Spoken Language Understanding}
Driven by recent advances in pretrained speech encoders, multilingual foundation models, and end-to-end architectures, modern Spoken Language Understanding (SLU) systems have increasingly moved beyond cascaded ASR and Natural Language Understanding (NLU) pipelines toward end-to-end models.
However, while high-resource languages benefit from large-scale annotated benchmarks and powerful pretrained models, Arabic SLU remains comparatively under-resourced, particularly for dialectal varieties.

\paragraph{MSA Models} Existing MSA SLU systems range from conventional statistical and neural approaches to recent pretrained Transformer-based models. Early Arabic SLU work relied on classical classifiers such as Multinomial Na\"{i}ve Bayes, Logistic Regression, Support Vector Machines, and CNNs for intent identification~\cite{lichouri2019arabic}.
For slot filling, earlier Arabic SLU systems have employed rule-based, statistical, and sequence-labeling approaches to identify semantic concepts and their relationships within user utterances~\cite{lhioui2013combined}.
A major resource for evaluating speech-based MSA SLU is SpeechMASSIVE~\cite{lee24i_interspeech}, which extends the MASSIVE benchmark~\cite{fitzgerald2023massive} to the speech modality and includes Arabic among its supported languages. The benchmark covers 18 domains, 60 intents, and 55 slot types, enabling the evaluation of both intent classification and slot filling from speech. However, the Arabic subset remains extremely limited, with only 115 training utterances, making it more suitable for few-shot and cross-lingual evaluation than for training dedicated MSA SLU models. This scarcity of annotated speech-semantic data remains one of the main challenges for advancing MSA SLU.

\paragraph{DA Models} Recent work on dialectal Arabic SLU has mainly focused on low-resource Tunisian Arabic.
\cite{mdhaffar2024performance} benchmarked a range of pretrained speech encoders for end-to-end SLU on TARIC-SLU~\cite{mdhaffar2024taric}, including monolingual models such as wav2vec~2.0~\cite{baevski2020wav2vec}, HuBERT~\cite{hsu2021hubert}, WavLM~\cite{chen2022wavlm}, and data2vec~2.0~\cite{baevski2023efficient}, as well as multilingual models such as XLS-R~\cite{babu22_interspeech}, MMS~\cite{pratap2024scaling}, MMS-1B~\cite{pratap2024scaling}, and w2v-BERT~2.0~\cite{barrault2023seamless}.
Among the evaluated systems, w2v-BERT~2.0 achieved the best overall SLU performance. More recently, semantically enhanced models such as SENSE~\citep{mdhaffar2025sense}, which incorporate textual semantic information through a teacher--student training strategy, have been investigated for Tunisian SLU and have demonstrated further performance improvements.
More recently, SLURP-TN~\cite{slurptn} was introduced as a new resource for Tunisian Arabic SLU, providing speech-level intent and slot annotations for task-oriented conversational scenarios.


\begin{table*}[!ht]
\centering
\scriptsize
\setlength{\tabcolsep}{2.5pt}
\renewcommand{\arraystretch}{1.08}

\begin{tabular}{
@{}
>{\raggedright\arraybackslash}p{3.65cm}
>{\centering\arraybackslash}p{1.10cm}
>{\centering\arraybackslash}p{1.10cm}
>{\centering\arraybackslash}p{1.15cm}
>{\centering\arraybackslash}p{1.15cm}
>{\centering\arraybackslash}p{1.40cm}
>{\columncolor{gray!10}\centering\arraybackslash}p{1.70cm}
>{\columncolor{gray!20}[\tabcolsep][0pt]\centering\arraybackslash}p{3.30cm}
@{}
}

\toprule
\textbf{\mbox{Comparison Dimension}}
& \textbf{2020}
& \textbf{2021}
& \textbf{2022}
& \textbf{2023}
& \textbf{2024}
& \textbf{2025}
& \textbf{2026} \\

\midrule
\multicolumn{8}{c}{\textbf{Edition Overview}} \\

Modality
& Text
& Text
& Text
& Text
& Text
& Speech
& Speech \\

\# Subtasks
& 2
& 4
& 2
& 3
& 3
& 3
& 8 \\

Task families
& T1
& T1
& T1, T2
& T1, T3
& T1, T3, T4
& T1, T5, T6
& T1, T5, T7--T9 \\

\midrule
\multicolumn{8}{c}{\textbf{Task Coverage}} \\

Dialect Identification (T1)
& \checkmark
& \checkmark
& \checkmark
& \checkmark
& \checkmark
& \checkmark
& \checkmark \\

Sentiment Analysis (T2)
& --
& --
& \checkmark
& --
& --
& --
& -- \\

Dialect-to-MSA MT (T3)
& --
& --
& --
& \checkmark
& \checkmark
& --
& -- \\

ALDi Estimation (T4)
& --
& --
& --
& --
& \checkmark
& --
& -- \\

Automatic Speech Recognition (T5)
& --
& --
& --
& --
& --
& \checkmark
& \checkmark \\

Diacritic Restoration (T6)
& --
& --
& --
& --
& --
& \checkmark
& -- \\

Text-to-Speech (T7)
& --
& --
& --
& --
& --
& --
& \checkmark \\

Spoken Language Translation (T8)
& --
& --
& --
& --
& --
& --
& \checkmark \\

Spoken Language Understanding (T9)
& --
& --
& --
& --
& --
& --
& \checkmark \\

\midrule
\multicolumn{8}{c}{\textbf{Task Characteristics}} \\

Province-level DID
& \checkmark
& \checkmark
& --
& --
& --
& --
& -- \\

MSA/DA Geographic DID
& --
& \checkmark
& --
& --
& --
& --
& -- \\

Multi-label DID
& --
& --
& --
& --
& \checkmark
& --
& -- \\

Robust / Low-bandwidth ASR
& --
& --
& --
& --
& --
& --
& \checkmark \\

Mixed-dialect ASR
& --
& --
& --
& --
& --
& --
& \checkmark \\

Code-switched ASR
& --
& --
& --
& --
& --
& --
& \checkmark \\

Out-of-domain SDID
& --
& --
& --
& --
& --
& --
& \checkmark \\

Zero-shot Domain Evaluation
& --
& --
& --
& --
& --
& --
& \checkmark \\

Open/closed Tracks
& --
& --
& --
& C/O
& C/O
& C/O (T6)
& O (T7) \\

\midrule
\multicolumn{8}{c}{\textbf{Geographic / Variety Scope}} \\

DI Coverage
& 21 countries
& 21 countries
& 18 countries
& 18 countries
& Country-level
& 8 dialects
& \shortstack{6 countries\\18 city dialects} \\

Province-level Coverage
& 100 provinces
& 100 provinces
& --
& --
& --
& --
& -- \\

MT / SLT Coverage
& --
& --
& --
& 4 dialects
& 4 dialects
& --
& 8 dialects \\

TTS Coverage
& --
& --
& --
& --
& --
& --
& 10 countries \\

SLU Coverage
& --
& --
& --
& --
& --
& --
& Tunisian \\

Robust ASR Coverage
& --
& --
& --
& --
& --
& --
& \shortstack{
Low-bandwidth /\\
degraded speech
} \\

Mixed-dialect ASR Coverage
& --
& --
& --
& --
& --
& --
& \shortstack{
6 countries\\
$>$15 subdialects
} \\

Code-switched ASR Coverage
& --
& --
& --
& --
& --
& --
& \shortstack{
Tunisian +\\
EN/FR
} \\

\midrule
\multicolumn{8}{c}{\textbf{Evaluation}} \\

Metric families
& M1
& M1
& M1, M2
& M1, M3
& M1, M3, M4
& M5--M8
& M3, M5--M14 \\

\midrule
\multicolumn{8}{c}{\textbf{Participation and Impact}} \\

Registered teams
& 61
& 53
& 41
& 58
& 51
& 44
& 65 \\

Participating teams
& 18
& 8
& 21
& 18
& 12
& 8
& 21 \\

Test submissions
& 56
& 68
& 105
& 76
& 76
& 100
& 48 \\

Accepted papers
& 14
& 7
& 15
& 13
& 8
& 7
& 14 \\


\bottomrule
\end{tabular}

\caption{
Detailed evolution of the NADI shared task series from 2020 to 2026.
The series focused on text-based tasks from 2020 to 2024, transitioned
to speech in 2025, and further expanded its speech-processing scope in 2026.
Task-family codes are normalized across editions and do not correspond
to the original task numbering within individual NADI editions.
T1: Dialect Identification;
T2: Sentiment Analysis;
T3: Dialect-to-MSA Machine Translation;
T4: Arabic Level of Dialectness Estimation;
T5: Automatic Speech Recognition;
T6: Diacritic Restoration;
T7: Text-to-Speech;
T8: Spoken Language Translation; and
T9: Spoken Language Understanding.
Metric-family codes are normalized across editions.
M1: Macro-F1;
M2: F1-PN;
M3: BLEU;
M4: Root Mean Squared Error;
M5: Accuracy;
M6: Average Cost (Cavg);
M7: Word Error Rate;
M8: Character Error Rate;
M9: UTMOSv2;
M10: NISQA;
M11: chrF;
M12: Weighted F1;
M13: Concept Error Rate;
M14: Concept/Value Error Rate.
DI: Dialect Identification;
MSA: Modern Standard Arabic;
DA: Dialectal Arabic;
SDI: Spoken Dialect Identification;
ASR: Automatic Speech Recognition;
MT: Machine Translation;
TTS: Text-to-Speech;
SLT: Spoken Language Translation;
SLU: Spoken Language Understanding;
ALDi: Arabic Level of Dialectness;
C/O: Closed/Open;
EN: English;
FR: French.
For task-property rows, a check mark indicates that the property is present,
whereas a dash (--) indicates that it is absent.
For numeric rows, a dash indicates unavailable or unreported information.
}
\label{tab:nadi_evolution_full}

\end{table*}


\begin{table*}[!ht]
\centering
\small 
\begin{tabular}{lccc}
\toprule
 \textbf{Subtask}  & Codabench Participants & Dev Submissions & Test Submissions  \\
\midrule
1.1 Robust ASR & 30 & 5 & 9  \\
1.2 Multidialect ASR & 19 & 2 & 4  \\
1.3 CS ASR & 17 & 4 &  6 \\
2 SDID & 39 & 6  & 14$^*$  \\
3 TTS & N/A & N/A & 1 \\
4 SLT & 15 & 3 & 2 \\
5.1 Intent Classification & 16 & 2 &  8$^\dagger$ \\
5.2 Slot Filling & 8 & 0 & 4$^\ddag$  \\

\bottomrule

\end{tabular}
\caption{Overview of participation across the eight subtasks. We count the number of people signed up for each Codabench and number of teams that publicly submitted for both the Dev and Test phase. Test submissions counts include: $*$ two duplicate submissions; $\dagger$ one unregistered and two duplicate submissions; $\ddag$ two invalid submissions. For the final leaderboard only the top performing submission for each team is included.}
\label{tab:participation}
\end{table*}

\section{Additional Task Information}
\label{sec:appendix_task_supp}

\subsection{TTS Data Overview}
\label{sec:app_tts_data}
Following extraction, the text underwent a structured preprocessing pipeline. We normalized Unicode representations (NFKC), removed diacritics, Tatweel, emojis, parenthetical metadata, HTML tags, and non-Arabic characters. We further normalized character runs (reducing repeated characters and removing online laughter), stripped non-standard punctuation, and enforced strict quality constraints by removing sentences containing English words, URLs, user tags, long gibberish words, or fewer than 5 words. In total, the benchmark comprises 4,001 samples spanning the 10 target dialects, sourced from public datasets, web data, and private team resources. The country-wise distribution of the evaluation samples is detailed in Table~\ref{tab:country_samples}.

\subsection{SLT Data Overview}
We provide an overview of the duration and word counts for the SLT subtask in Table \ref{tab:whitehouse_test}.
\begin{table}[h]
\small 
    \centering
 
    \resizebox{\columnwidth}{!}{
    \begin{tabular}{llr}
        \toprule
        \textbf{Country Code} & \textbf{Country Name} & \textbf{Sample Count} \\
        \midrule
        \texttt{ae} & United Arab Emirates & 537 \\
        \texttt{dz} & Algeria              & 293 \\
        \texttt{eg} & Egypt                & 424 \\
        \texttt{jo} & Jordan               & 631 \\
        \texttt{ma} & Morocco              & 201 \\
        \texttt{ps} & Palestine            & 750 \\
        \texttt{sa} & Saudi Arabia         & 100 \\
        \texttt{sd} & Sudan                & 114 \\
        \texttt{tn} & Tunisia              & 623 \\
        \texttt{ye} & Yemen                & 328 \\
        \midrule
        \textbf{Total} &                    & \textbf{4,001} \\
        \bottomrule
    \end{tabular}

    }\caption{Distribution of evaluation text samples across target dialects. }    \label{tab:country_samples}

\end{table}

\begin{table}[h]
\small 
    \centering

    {
    \begin{tabular}{llr}
        \toprule
        \textbf{Dialect} & \textbf{Duration} & \textbf{\# words (AR/EN)} \\
        \midrule
         Algeria     & 0:55:03 & 8,032/10,559\\
         Egypt       & 0:59:12 & 10,116/12,228\\
         Jordan      & 0:58:54 & 8,804/12,188\\
         Mauritania  & 0:56:15 & 10,116/12,228  \\ 
         Morocco     & 1:00:23 & 11,959/15,796\\
         Palestine   & 0:58:22 & 8,833/11,729\\
         UAE         & 0:55:59 & 8,766/12,163 \\
         Yemen       & 1:02:30 & 9,359/13,296 \\
    \textbf{Total}   & \textbf{7:46:38} & \textbf{75,985/ 100,187} \\
        \bottomrule
    \end{tabular}
    }    \caption{Duration (h:mm:ss) and number of words
per dialect of the NADI 2026 official speech translation evaluation set. \# words represent the number of words
per dialect in the original dialectal transcription (AR)
and the corresponding English translation (EN).}
    \label{tab:whitehouse_test}
\end{table}

\subsection{SLU Data Overview}
\label{sec:app_slu}
The evaluation material was recorded by 12 speakers between 23 and 28 years of age.
The speaker distribution is exactly balanced, with six male and six female speakers. At the
utterance level, male speakers contributed 523 utterances and female speakers contributed
466, corresponding to 52.88\% and 47.12\% of the evaluation set, respectively.
The speakers represent five regional backgrounds. Five speakers are associated with
Greater Tunis, four with Medenine, and one each with Kebili, Kairouan, and Sfax. These
speakers contributed 492, 322, 110, 41, and 24 utterances, respectively. Although the regional
distribution is not uniform, it introduces variation beyond a single urban variety and
preserves the broader regional motivation of SLURP-TN.
A central design choice was to combine domains already represented in SLURP-TN with
previously unseen domains. Emails, Weather, and General constitute the three returning
domains.
Calendar, Question Answering, Music, and Cooking constitute four new domains.
These domains are absent from the SLURP-TN
training and development sets provided to participants. They therefore support zero-shot
evaluation at the domain level: systems must process utterances belonging to application
domains that were not represented during system development.
The official evaluation set consequently assesses two complementary forms of generalisation.
For the returning domains, systems must generalise to new speakers and previously
unseen utterances while operating within familiar application domains. For the zero-shot
domains, systems must additionally transfer their learned acoustic, lexical, and semantic
representations to new application areas.
\section{Baseline Systems}
\label{sec:appendix_baselines}

\paragraph{Task 1 ASR:}

For all three of the ASR subtasks we use \texttt{whisper-Large-v3}~\cite{radford2023robust}, a multilingual ASR model trained on a noisy-labeled internet audio. During the models training, there was no distinguishing between varieties of Arabic, however, it can be assumed that by training off of subtitles, there may be a preference for generating MSA. 

\paragraph{Task 2 Spoken Dialect Identification:} 
We adopt the baseline from NADI 2025 ADI subtask~\cite{talafha-etal-2025-nadi}, with a number of modifications. It starts with a ECAPA-TDNN system~\footnote{\url{https://huggingface.co/speechbrain/lang-id-voxlingua107-ecapa}} originally trained on LinguaVox107~\cite{valk2021voxlingua107}, which we then finetune on a 10hr-per-dialect version of ADI-20 dataset. 

We use a learning rate schedule as follows starting with only the final classification layer thawed: 9k steps linear warm up from $3.3*10^{-5}$ to $10^{-4}$, 6k steps at constant LR, thaw remaining layers, 15k steps linear warm up from $3.3*10^{-5}$ to $10^{-4}$, 30k steps constant LR, and a final 30k steps with exponential decay. \\

\paragraph{Task 3 Text-To-Speech:} 
We used the XTTS-v2 model as the baseline for this subtask to generate audio samples across all dialects. A single reference speaker, ``Suad Qasim,'' was selected and used consistently across all dialects. XTTS is a Cross-Lingual TTS model developed by Coqui~\footnote{\url{https://docs.coqui.ai/en/latest/index.html}} that can synthesize speech from text while using a short reference audio clip to reproduce characteristics of a target speaker’s voice. While this baseline was not available for comparison during the original evaluation period, we provide it to better contextualize the submission we received.\\

\noindent
\textbf{Task 4 Spoken Language Translation:} 
We use \texttt{whisper-large-v3} as a baseline model to generate the English translation
of all the considered Arabic dialects. Whisper is an end-to-end multi-task speech model that adopts a transformer-like encoder-decoder architecture. It natively handles language identification, speech transcription, and speech translation directly into English text.\\

\noindent
\textbf{Task 5 Spoken Language Understanding:} 
We use \texttt{whisper-small} as the backbone for both of Task 5's baselines. For the intent recognition subtask, we simply add masked average pooling on the encoder output and pass this to a linear layer for classification. For slot filling, we directly finetune \texttt{whisper-small} to add the slot labeling as part of the output transcription. In both parts we use SLURP-TN for training data.
\section{Additional Results}
\label{sec:appendix_results}
Due to space constraints we provide tables of results for the remaining subtasks:
\begin{itemize}

    \item Subtask 1.3 Code-switched ASR: Table~\ref{tab:nadi26-subtask13-systems}.
    \item Subtask 2 SDID: Table~\ref{tab:adi-2-res}.
    \item Subtask 3 TTS: Table~\ref{tab:tts-task3-results}.
    \item Subtask 4 SLT: Table~\ref{tab:ast-task4-results}
    \item Subtask 5.1 Intent Classification: Table~\ref{tab:slu_5_1_res}
    \item Subtask 5.2 Slot Fillings: Table~\ref{tab:slu-5-2-res}
\end{itemize}

\begin{table*}[t]
\centering
\setlength{\tabcolsep}{3pt}
\small 
\begin{tabular}{@{}lrrrrlllcl@{}}
\toprule
&  \multicolumn{2}{c}{\textbf{Official}} & \multicolumn{2}{c}{\textbf{Hamza-norm.}} & & & & & \\
\cmidrule(lr){2-3} \cmidrule(lr){4-5}
\textbf{Team} & \textbf{WER}$\downarrow$  & \textbf{CER}$\downarrow$ & \textbf{WER}$\downarrow$ & \textbf{CER}$\downarrow$ &
\textbf{Backbone} & \textbf{Adaptation} & \textbf{Data+} &
\textbf{CS-targeted} & \textbf{Ensemble} \\
\midrule
Abjad AI     & \textbf{14.41} & 5.55 & \textbf{14.12} & 5.43 & Cohere & Full FT & $\checkmark$      & $\checkmark$ & $\checkmark$(5)  \\
Itgan         & 14.49 & \textbf{5.38} & 14.22 & \textbf{5.27} & \texttt{whisper-large-v3}              & LoRA    &                               &            &  $\checkmark$(3) \\
Thakaa          & 14.53 & 6.34 & 14.26 & 6.26 & Heterogeneous     & Full FT & $\checkmark$    &            & $\checkmark$(10) \\
KAND CA         & 15.21 & 5.86 & 14.91 & 5.73 & \texttt{whisper-large-v3}$*$   & Full FT &$\checkmark$   & $\checkmark$ &  \\
NAMAA  & 16.98 & 6.58 & 16.68 & 6.46 & \texttt{FarukSTT} (Whisper)            & LoRA    &                               & $\checkmark$ &  \\
Ar\_India        & 73.76 & 49.56 & 71.81 & 48.91 & \cellcolor[gray]{0.8}           & \cellcolor[gray]{0.8} & \cellcolor[gray]{0.8}  &     \cellcolor[gray]{0.8}   & \cellcolor[gray]{0.8}\\
Baseline  & 88.60 & 66.14 & 87.91 & 65.61 & \texttt{whisper-large-v3}           &   &                               & &  \\

\bottomrule
\end{tabular}%
\caption{Official Subtask~1.3 (Code-switched ASR) results and reported system characteristics for
the six teams that provided system descriptions, ordered by percentage WER; percentage CER and the hamza-normalized variants of both
metrics are also shown, and lower is better. \textit{Cohere} refers to the \texttt{cohere-transcribe-arabic-07-2026} model. $*$ Finetuned dialectal checkpoint. \textbf{Adaptation}:
distinguishes full fine-tuning from LoRA; \textbf{Data+}: whether additional data was used on top of the
Subtask~1.3 releases, excluding checkpoint pretraining; \textbf{CS-targeted}:
marks measures aimed specifically at code-switched segments, such as adding
non-Arabic speech, oversampling Latin-script references, or lifting Arabic-only
decoding constraints; and \textbf{Ensemble}: indicates the number of components used in the final ensemble. All teams except \textsc{Thakaa} 
reported acoustic augmentation and used an external decoding language
model, and \textsc{KAND CA} alone applied output post-processing. \textsc{ar\_india} did not submit a system paper. }
\label{tab:nadi26-subtask13-systems}
\end{table*}

\begin{table*}[!ht]
\centering
\small
\begin{tabular}{lccccccc}
\toprule
 & \textbf{Accuracy} $\uparrow$ & \textbf{C}$_{avg}\downarrow$ & \textbf{Audio} & \textbf{Text Branch} & \textbf{Ensemble} & \textbf{Augm.} & \textbf{OOD} \\
\midrule
Thakaa	&\textbf{56.15}	&11.31 & Ara-BEST-RQ &$\checkmark$ & $\checkmark$(5) &   & \\
Salesteq&	54.67	&\textbf{8.64} & \cellcolor[gray]{0.8} & \cellcolor[gray]{0.8} & \cellcolor[gray]{0.8} & \cellcolor[gray]{0.8} & \cellcolor[gray]{0.8}   \\
Lynx	&53.53	&12.47 & Cohere  & $\checkmark$ &    &      & $\checkmark$ \\
AI Elites& 	52.51	&11.39 & Cohere & &    & $\checkmark$ & $\checkmark$\\
KAND CA&	51.14	&13.26 & Cohere & $\checkmark$ & $\checkmark$(2) &  &   \\
Abjad Ai&	41.57	&14.73 & Cohere & &   & $\checkmark$ &\\
Resonate&	40.89	&16.78 & WavLM & &   & $\checkmark$ &\\
FARABI-Fraunhofer-IAIS	&39.18&	16.95 &\cellcolor[gray]{0.8}   & \cellcolor[gray]{0.8}     &  \cellcolor[gray]{0.8}  & \cellcolor[gray]{0.8}  & \cellcolor[gray]{0.8}  \\
CodeZone Research group&	39.18&	14.69 & SB+Whisper & & $\checkmark$(2) & &\\
Namaa &	37.70&	16.74 & SB+Whisper &  & $\checkmark$(4) &    &\\
Ar\_India	&31.21&	20.26 & \cellcolor[gray]{0.8}  & \cellcolor[gray]{0.8}  & \cellcolor[gray]{0.8} & \cellcolor[gray]{0.8} & \cellcolor[gray]{0.8}\\
ArabSpeech Minds&	30.64&	16.97 & \cellcolor[gray]{0.8}  &\cellcolor[gray]{0.8} & \cellcolor[gray]{0.8} & \cellcolor[gray]{0.8} & \cellcolor[gray]{0.8}\\
Baseline &	28.70 &	20.02 & SB & &  & & \\ \hline
\end{tabular}
\caption{Results from Subtask 2 (Spoken Dialect Identification) alongside submission approach. Best results in bold. We report $C_{avg}*100$ for readability.  \textbf{Audio}: audio models used with shorthands for \textit{Cohere} (\texttt{cohere-transcribe-arabic-07-2026}) \textit{SB} (Speech Brain \texttt{lang-id-voxlingua107-ecapa}), and \textit{Whisper} for any of the Whisper family models; \textbf{Text Branch}: if the approach uses a multimodal text+speech pipeline; \textbf{Ensemble}: number of ensemble component models;  \textbf{Augm.}: if any data augmentation approach was used; \textbf{OOD}: if teams applied specific methods to enhance out-of-domain performance.
\textsc{Salesteq}, \textsc{FARABI-Fraunhofer}, \textsc{ar\_india}, and \textsc{ArabSpeech Minds} did not submit system papers. }
\label{tab:adi-2-res}
\end{table*}



\begin{table*}[t]
\centering
\small
\setlength{\tabcolsep}{8pt}
\renewcommand{\arraystretch}{1.10}

\begin{tabular}{llr|lr|lr|lr|lr}
\toprule
\textbf{Dialect / Country} &
\multicolumn{2}{c}{\textbf{UTMOS}} $\uparrow$ &
\multicolumn{2}{c}{\textbf{NISQA-MOS}} $\uparrow$ &
\multicolumn{2}{c}{\textbf{NISQA-TTS}} $\uparrow$ &
\multicolumn{2}{c}{\textbf{WER}} $\downarrow$ &
\multicolumn{2}{c}{\textbf{CER}} $\downarrow$ \\
& \textbf{K} & \textbf{B} & \textbf{K} & \textbf{B} & \textbf{K} & \textbf{B} & \textbf{K} & \textbf{B} & \textbf{K} & \textbf{B} \\
\midrule

United Arab Emirates  & 2.73 & \textbf{3.10} & 3.75 & \textbf{4.29} & 3.45 & \textbf{3.85} & \textbf{31.31} & 31.71 & 15.45 & \textbf{8.45}\\
Algeria               & 2.60 & \textbf{3.13}  & 3.81 & \textbf{4.31} & 2.96 & \textbf{3.95} & 44.63 & \textbf{40.82} & 13.01 & \textbf{10.13} \\
Egypt                 & 2.81 & \textbf{3.15} & 2.75 & \textbf{4.30} & 3.75 & \textbf{3.94} & \textbf{20.49} & 28.59 & \textbf{5.37} & 7.33  \\
Jordan                & 2.60 & \textbf{3.10} & 4.21 & \textbf{4.29} & 3.18 & \textbf{3.91} & 32.84 & \textbf{30.03} & 8.19 & \textbf{6.98}  \\
Morocco               & \textbf{3.23} & 3.13 & 3.52 & \textbf{4.39} & 3.05 & \textbf{3.87} & 55.46 & \textbf{27.34} & 21.62 & \textbf{6.17} \\
Palestine             & 2.58 & \textbf{3.10} & 4.23 & \textbf{4.38} & 3.09 & \textbf{3.90} & \textbf{20.16} & 20.33 & 5.66 & \textbf{5.33} \\
Saudi Arabia          & 2.79 & \textbf{3.09} & 3.85 & \textbf{4.28} & 3.45 & \textbf{3.84} & 14.28 & \textbf{13.59} & 4.87 & \textbf{3.57} \\

Sudan                 & \textbf{3.19} & 3.10 & \textbf{4.46} & 4.21 &  \textbf{3.98} & 3.96 & 35.71 & \textbf{29.98} & 9.96 & \textbf{7.52}  \\
Tunisia               & 3.04 & \textbf{3.13} & 2.74 & \textbf{4.30} & 3.20 & \textbf{3.94} & 44.52 & \textbf{34.18} & 12.23 & \textbf{7.75} \\
Yemen                 & 2.17 & \textbf{3.09} & 2.65 & \textbf{4.29} & 2.46 & \textbf{3.87} & 47.92 & \textbf{31.31} & 25.12 & \textbf{7.97} \\
\midrule
\textbf{Average}   & 2.72 & \textbf{3.11}   & 3.57 & \textbf{4.31} & 3.21 & \textbf{3.91}  & 32.48 & \textbf{29.28} & 11.40 & \textbf{7.22} \\

\bottomrule
\end{tabular}

\caption{Detailed dialect-level evaluation of the \textsc{KAND CA} submission (column \textbf{K}) for
Task~3 (TTS), against the \textsc{XTTS-v2} baseline (column \textbf{B}). UTMOS, NISQA-MOS, and NISQA-TTS evaluate
speech quality, with higher scores indicating better performance, while
WER and CER (percentages) measure transcription errors, with lower values indicating
better intelligibility. Bold values indicate the best performance.}
\label{tab:tts-task3-results}
\end{table*}

\begin{table*}[t]
\centering
\small
\setlength{\tabcolsep}{5pt}
\renewcommand{\arraystretch}{1.12}

\begin{tabular*}{\textwidth}{
@{\extracolsep{\fill}}
lcccccccccc
@{}
}
\toprule
&
\multicolumn{2}{c}{\textbf{Overall}} &
\multicolumn{8}{c}{\textbf{Dialect BLEU}} \\
\cmidrule(lr){2-3}
\cmidrule(lr){4-11}

\textbf{Team} &
\textbf{BLEU} $\uparrow$ &
\textbf{chrF} $\uparrow$ &
\textbf{ALG} &
\textbf{EGY} &
\textbf{JOR} &
\textbf{MAU} &
\textbf{MOR} &
\textbf{PAL} &
\textbf{UAE} &
\textbf{YEM} \\
\midrule

\textsc{Sense} &
\textbf{14.19} &
\textbf{34.11} &
\textbf{12.16} &
\textbf{19.84} &
\textbf{27.46} &
\textbf{6.84} &
\textbf{8.51} &
\textbf{20.32} &
\textbf{0.39} &
\textbf{16.07} \\

\textsc{Abjad AI$\dagger$} &
5.77  &
22.94 &
7.08  &
0.14  &
20.65 &
3.53  &
0.21&
0.39  &
0.18  &
10.97  \\

Baseline &

   5.52 &
  20.28 &
   7.95 &
  0.06 &
  21.6 &
  3.22 &
   0.07 &
   0.35 &
  0.07 &
  11.0 \\

\bottomrule
\end{tabular*}

\caption{Evaluation results for Task~4 (Spoken Language Translation). Overall performance is reported using BLEU and chrF, while dialect-level performance is reported using BLEU; higher scores indicate better performance. Dialect abbreviations are ALG (Algeria), EGY (Egypt), JOR (Jordan), MAU (Mauritania), MOR (Morocco), PAL (Palestine), UAE (United Arab Emirates), and YEM (Yemen). Bold values indicate best performance. $\dagger$ This submission is a post-evaluation phase run which aimed to minimize an unintended use of development data in system training.}
\label{tab:ast-task4-results}
\end{table*}

\begin{table}[!th]
\centering
\small
\setlength{\tabcolsep}{6pt}
\renewcommand{\arraystretch}{1.10}

\begin{tabular}{clcc}
\toprule
\textbf{Rank} &
\textbf{Team} &
\textbf{F1} $\uparrow$ &
\textbf{Accuracy} $\uparrow$ \\
\midrule

1 & \textbf{Lynx}             & \textbf{76.86} & \textbf{75.73} \\
2 & CIS-OpenSLU               & 71.54          & 72.50 \\
3 & A3da2 Messi                & 69.92 & 70.07 \\
4 & Aslema                    & 66.93          & 66.13 \\
5 & FARABI-Fraunhofer         & 64.69          & 65.93 \\
6 & Baseline                 & 11.08  & 13.35 \\
\bottomrule
\end{tabular}

\caption{Official results for Task~5.1 (Intent Recognition). Systems are ranked by weighted F1; higher weighted F1 and accuracy indicate better performance. }
\label{tab:slu_5_1_res}
\end{table}

\begin{table}[t]
\centering
\small
\setlength{\tabcolsep}{10pt}
\renewcommand{\arraystretch}{1.10}

\begin{tabular}{clcc}
\toprule
\textbf{Rank} &
\textbf{Team} &
\textbf{CoER} $\downarrow$ &
\textbf{CVER} $\downarrow$ \\
\midrule

1 & \textbf{Aslema} & \textbf{59.53} & \textbf{94.20} \\
2 & Zila            & 72.78          & 100.36 \\
3 & Baseline & 100.36 & 120.95 \\

\bottomrule
\end{tabular}

\caption{Official results for Task~5.2 (Slot Filling). Systems are ranked by Concept Error Rate (CoER); lower CoER and Concept-Value Error Rate (CVER) indicate better performance. 
}
\label{tab:slu-5-2-res}
\end{table}
\section{Complete Submission Descriptions}
\label{sec:appendix_systems}

\subsection{Subtask 1.1: Robust Dialectal ASR}

\paragraph{Wifaq (Rank 1)~\cite{wifaq-nadi2026}.}
Wifaq fine-tuned four character-level connectionist temporal classification
(CTC) recognizers~\cite{graves2006connectionist} from \texttt{w2v-bert-2.0} using different training stages and
mixtures of NADI, Casablanca, selected Omnilingual ASR subsets~\cite{omnilingual2025omnilingual}, and MoulSot
Moroccan Arabic~\cite{moulsot2026}. Each model used beam search with a five-gram KenLM trained on
speech transcripts and approximately 440,000 dialectal tweets; word-level
ROVER combined their hypotheses.

\paragraph{Resonate (Rank 2)~\cite{resonate-nadi2026}.}
Resonate fully fine-tuned the Arabic-specialized
\texttt{cohere-transcribe-arabic-07-2026} model on official data expanded
through neural denoising~\cite{zhao25f_interspeech}, $k$-nearest-neighbor voice conversion~\cite{baas23_interspeech},  and using a low-pass filter to better match audio energy characteristics. It conditioned the decoder on the
country label and used duration-aware SpecAugment for short clips~\cite{park2019specaugment}. Beam-search
output underwent dialect-specific orthographic canonicalization.

\paragraph{Thakaa (Rank 3)~\cite{thakaa-nadi2026}.}
Thakaa selected ROVER committees from seven candidates: one Cohere model, four
NeMo FastConformer~\cite{rekesh2023fast} variants, one Whisper LoRA model, and one Qwen3-ASR model~\cite{yang2025qwen3}.
Its strongest component was trained on Subtasks~1.1 and~1.3, while other
candidates used KenLM shallow fusion or LoRA adaptation. Validation-based
search selected one committee for five dialects and specialized committees for
Algeria, Mauritania, and Yemen.

\paragraph{KAND CA (Rank 4) ~\cite{kand-nadi2026}.}
The team continued full-model adaptation of \texttt{whisper-large-v3} on NADI
and Casablanca data with replay; its broader Arabic training mixture included
Lahgtna~\cite{kandca_lag}, Common Voice~\cite{ardila-etal-2020-common}, FLEURS~\cite{conneau2023fleurs}, MASC~\cite{al-fetyani2022masc}, and additional Algerian speech. Inference
used overlapping chunks with repetition removal. Eight dialect-specific
mT5-large spelling correctors were trained on synthetic corruptions of the
available official transcripts and applied to every hypothesis to better match
the task's orthographic conventions.

\paragraph{Ar\_India (Rank 5)
.}
The Ar\_India team did not submit a system paper.

\paragraph{Nile (Rank 6)~\cite{nile-nadi2026}.}
Nile used a two-stage pipeline based on \texttt{whisper-large-v3-turbo}; It first adapted the model on an
unspecified 200,000-recording corpus augmented with noise, reverberation, gain
changes, and background music, and then fine-tuned the checkpoint on the
official competition training set. This separated noise-robustness training
from task-specific dialect and transcription adaptation.

\paragraph{Itgan (Rank 7)~\cite{itgan-nadi2026}.}
Itgan trained rank-32 LoRA adapters using only the official data. A pooled
\texttt{whisper-large-v3-turbo} adapter was specialized for Algeria, Egypt,
Mauritania, and Morocco, while validation favored a non-Turbo Whisper backbone
for the other four countries. Country labels routed each utterance, and greedy
Arabic decoding included a hypothesis-length guard.

\paragraph{Sabaa (Rank 8)~\cite{sabaa-nadi2026}.}
Sabaa trained a separate rank-32 \texttt{whisper-large-v3-turbo} LoRA adapter
for each dialect using only official data; this outperformed a pooled adapter.
Beam-search decoding used beam size~8 and a negative length penalty, followed
by conservative dialect-specific spelling normalization.

\paragraph{Salesteq (Rank 9)
.}
The Salesteq team did not provide a system paper.

\subsection{Subtask 1.2: Mixed dialect ASR}
\paragraph{Salesteq (Rank 1)
}
The Salesteq team did not submit a system paper.

\paragraph{Namaa Community (Rank 2)~\cite{namaa-nadi2026}.}
They used \texttt{cohere-transcribe-arabic-07-2026} in a zero-shot setting, without task-specific fine-tuning. The Task~1.2 development set was used only for system selection, with no additional ASR training data included in the final configuration. Several alternatives were explored, including quantized inference, different decoding settings, orthographic normalization, and consensus-based hypothesis combination. The final system used full-precision inference without ensembling or additional text post-processing.

\paragraph{Itgan (Rank 3)~\cite{itgan-nadi2026}.}
The submission was based on \texttt{whisper-large-v3} and \texttt{whisper-large-v3-turbo}, with LoRA used for parameter-efficient adaptation. Different model variants were trained using distinct data configurations, and one component additionally incorporated 12.8K utterances from Task~1.1. No data external to NADI~2026 were used. The final system combined five recognizers using ROVER: three adapted \texttt{whisper-large-v3} variants, one adapted \texttt{whisper-large-v3-turbo} model, and one zero-shot \texttt{whisper-large-v3} model. Inference used the Faster-Whisper~\footnote{https://github.com/SYSTRAN/faster-whisper} interface for Whisper with beam size~5 and Silero VAD~\cite{SileroVAD}, with previous-text conditioning disabled.

\paragraph{KAND CA (Rank 4)~\cite{kand-nadi2026}.}
They used \texttt{whisper-large-v3} with sequential full-model fine-tuning from a checkpoint previously adapted to Arabic dialectal speech. During Task~1.2 adaptation, 1.5K Task~1.1 utterances and 1K Casablanca clips were replayed to preserve previously learned dialectal information, while earlier adaptation stages also incorporated additional Arabic dialect resources. The final system used a single model with chunked long-form decoding followed by repetition removal.

\subsection{Subtask 1.3: Code-switched Tunisian ASR}

\paragraph{Abjad AI (Rank 1)~\cite{abjad-nadi2026}.}
Abjad AI fully fine-tuned  \texttt{cohere-transcribe-arabic-07-2026}, keeping FP32 master weights under bfloat16 autocasting because pure bfloat16 rounded small AdamW updates to zero, and masking prompt positions from the loss. The official data was supplemented with text-to-speech-synthesized Tunisian speech, nine hours each of French and English read speech, and ESC-50 noise~\cite{piczak2015esc}, with synthetic recordings augmented more strongly than real ones; both online and offline waveform augmentation were used. The submission uniformly averaged five checkpoints trained under complementary data and augmentation configurations, averaging floating-point parameters in FP32 and retaining integer BatchNorm counters from one reference checkpoint.

\paragraph{Itgan (Rank 2)~\cite{itgan-nadi2026}.}
Itgan used only the shared-task corpus and trained rank-32 LoRA adapters on all attention and feed-forward projections of \texttt{whisper-large-v3}, 57.7\,M parameters or 3.6\% of the model, against targets processed by the organizer's cleaned-transcription form so that training matched exactly the text the metric scores. After a first run memorized heavily, raising LoRA dropout to 0.15 and enabling SpecAugment shrank the generalization gap by a factor of 3.5. Two combination stages followed with no further training: four adapters from two independent runs were averaged in weight space by concatenation, which yields the exact mean of the adapter updates as an ordinary higher-rank adapter, and then that model combination was used to pivot a three-system word-level ROVER vote that modified 63 of the 2,123 blind utterances. Doubling the LoRA rank, collapsing hamza variants and centroid hypothesis selection all failed to help.

\paragraph{Thakaa (Rank 3)~\cite{thakaa-nadi2026}.}
Thakaa submitted a ten-system word-level ROVER ensemble spanning CTC encoders (\texttt{w2v-BERT-2.0}, \texttt{mms-1b}), an encoder-decoder recognizer (\texttt{cohere-transcribe-arabic-07-2026}), an audio-LLM (\texttt{Qwen3-ASR-1.7B}) and two \texttt{whisper-large-v3} fine-tunes. Training added roughly 6,900 fully labeled TUNIFRA and Tunisian SLURP-derived utterances with no pseudo-labeling. A 4-gram KenLM gave shallow-fusion CTC decoding about 20\% relative WER over greedy decoding, while the autoregressive systems used beams of 5 to 10 with repetition constraints. Committee composition was searched over all checkpoints with bootstrap significance testing anchored on the strongest CTC system.

\paragraph{KAND CA (Rank 4)~\cite{kand-nadi2026}.}
KAND CA began from a broadly dialect-adapted \texttt{whisper-large-v3} checkpoint, which it measured as worth about 4.3 WER points over an English base after identical fine-tuning. Full fine-tuning with SpecAugment was followed by a single fresh two-epoch schedule over 70,595 clips combining roughly 46,000 external gold Tunisian-French clips with TEDx replay, reported as the decisive step; warm continuation and further epochs both regressed. Decoding was greedy, as beam search and $n$-gram blocking hurt, and post-processing was restricted to an accent-only filter that accepted a French spellchecker fix only when unaccenting it reproduced the original token, touching five tokens. Every learned corrector degraded results, most severely a \texttt{Qwen3.5-4B} LoRA corrector at 5.9 points.

\paragraph{NAMAA Community (Rank 5)~\cite{namaa-nadi2026}.}
NAMAA Community adapted FarukSTT~\footnote{https://huggingface.co/medyas/FarukSTT}, a Tunisian Whisper derivative, in two LoRA stages while using the \texttt{whisper-large-v3} processor so Arabic, French and English tokens could co-occur. Stage~1 trained rank-32 adapters on the query and value projections and merged them; stage~2 started from that merge, added the publicly released legacy test split, and duplicated every Latin-script example to increase code-switch exposure, giving 38,106 examples from 25,730 sources. Adapters covered the attention projections, feed-forward layers and output projection, with checkpoint selection on validation WER. Inference used beam size~3 with forced decoder IDs and token suppression disabled to lift the Arabic-script restriction, and raw transcriptions were submitted because the scorer normalizes both sides.

\paragraph{Ar\_India (Rank 6)
}
The Ar\_India team did not submit a system paper

\subsection{Subtask 2: Spoken Dialect ID}

\paragraph{Thakaa (Rank 1)~\cite{thakaa-nadi2026}.} This multimodal system use Ara-BEST-RQ~\cite{elleuch2026ara} for its audio component alongside multiple text pipelines. Two text pipelines were trained, one using standard transcriptions, and the second stream using fused transcriptions, auxiliary branch loss and transcription-view dropout. Each of the pipelines consisted of a Cohere Arabic transcription branch and a \texttt{whisper-large-v3} branch. Training consisted of a mix of open resources totally 265 hours from eight distinct ADI and ASR datasets.

\paragraph{Salesteq (Rank 2)
}
The Salesteq team did not submit a system paper.

\paragraph{Lynx (Rank 3)~\cite{lynx-nadi2026}.}
Team Lynx uses \texttt{cohere-transcribe-arabic-07-2026} as a multimodal backbone with a text-branch utilizing TF-IDF features of the transcripts directly, while the acoustic branch uses the Cohere model as a frozen feature extractor before classifying the utterances with a two layer LSTM model. Their pooling method for the acoustic branch is of note, and involves temporal mean pooling first before passing all layers as a sequence to their LSTM classifier.

\paragraph{AI Elites (Rank 4)~\cite{ai_elites-nadi2026}.}
The AI Elites team focused on an speech-only model, taking \texttt{cohere-transcribe-arabic-07-2026} and finetuning it for SDID on the micro ADI20 dataset. Modifications to the model include using a learnable weighted sum through all encoder layers. Data augmentation was comprehensive and included noise and codec compression; speech and gain permutation; as well as SpecAugment~\cite{park2019specaugment} and random cropping. Key to their approach is using out-of-domain dataset for checkpoint selection, of which they used both Casablanca~\cite{talafha2024casablanca} and the MADIS-5~\cite{abdullah25_interspeech} as a secondary check.

\paragraph{KAND CA (Rank 5)~\cite{kand-nadi2026}.}
This multimodal submission used an ensemble, the first part of which was a finetuned FastConformer using an attention pooling head on top of \texttt{cohere-transcribe-arabic-07-2026}. To this they included the same classifier applied to a LavaSR~\footnote{https://github.com/ysharma3501/LavaSR} speech-enhanced version of the target audio, weighted evenly with the non-enhanced audio. Finally they included a text route classifier weighted at 30\%, which consisted of a finetuned \texttt{whisper-large-v3} model trained on the 1.1 subtask, fed to a BERT based dialect classifier.

\paragraph{Abjad Ai (Rank 6)~\cite{abjad-nadi2026}.}
This team also used \texttt{cohere-transcribe-arabic-07-2026} as starting point, and finetuned on the micro ADI20 dataset using a simple two layer classifier head with GELU activation and dropout on top of the fully frozen conformer model. 
For data augmentation, they applied a mix of methods including gain, additive noise, speed perturbation, and time-shifting. To reduce channel-sensitivity normalize loudness as well as apply a simulated bandwidth reduction.

\paragraph{Resonate (Rank 7)~\cite{resonate-nadi2026}.}
This team uses \texttt{wavlm-base-plus}~\cite{chen2022wavlm} as a base model with output passed to context-aware multi-head factorized attentive pooling to a final linear layer for classification. Models were trained with additive-angle margin softmax. A 100hr per dialect subset of ADI17 was used alongside the new additions from ADI20. Data filteration included removing short clips or clips with little speech as determined by a VAD. All training utterances were cut down to five seconds for training, meaning longer samples could lead to more distinct clips for training. The starting point of these samples were altered every few epochs. Data augmentation included speed, reverb, additive noise, frequence and time dropping.

\paragraph{FARABI-Fraunhofer  (Rank 8)
} 
The FARABI-Fraunhofer  team did not submit a system paper.

\paragraph{CodeZone Research group (Rank 9)~\cite{codezone-nadi2026}.}
This group uses an ensemble of speech models including a pretrained VoxLingua107 ECAPA-TDNN system~\footnote{https://huggingface.co/speechbrain/lang-id-voxlingua107-ecapa} and the pretrained ADI20 Whisper model from Elyadata~\footnote{https://huggingface.co/Elyadata/ADI-whisper-ADI20}. Both models are finetuned on the ADI20-10hr using a mix of data augmentation methods.  

\paragraph{Namaa Community (Rank 10)~\cite{namaa-nadi2026}.} This four component ensemble approach focused on models trained on the ADI20 dataset. The base models used included a VoxLingua107 pretrained ECAPA-TDNN system and finetuned with on the ADI20-10hr split; \texttt{whisper-large-v3} finetuned with LoRA on the ADI20-10hr split; \texttt{whisper-large-v3} trained on the ADI20-53hr split; and finally a \texttt{whisper-large-v3} model finetuned on the ADI17 dataset. Logits were combined using a weighted sum approach picked using the development set.

\paragraph{Ar\_India (Rank 11)
} 
The Ar\_India team did not submit a system paper.

\paragraph{Arabspeech Minds (Rank 12)
}
The Arabspeech Minds team did not submit a system paper.

\subsection{Task 4 SLT}

\paragraph{Sense (Rank 1)
} 
Team Sense did not submit a system paper.

\paragraph{Abjad AI (Rank 2)~\cite{abjad-nadi2026}.}
The system used \texttt{whisper-large-v3} with full-model fine-tuning for Arabic-to-English speech translation. Two adaptation strategies were explored: intermediate fine-tuning on an internal out-of-domain Arabic speech corpus followed by task-specific adaptation, and direct fine-tuning on the shared-task data. The latter yielded better development performance and was adopted for the final system. To address the limited amount of training data, the speech was augmented using Gaussian noise, time stretching, pitch shifting, and temporal shifting, and the training set was upsampled to approximately \(2.5\) hours. The final model was fine-tuned directly on the augmented and upsampled Task~4 data. During a review of the system, the team discovered some of their training data used was shown to overlap with the development dataset, the reported results are thus a re-run of their system with this overlap minimized.

\subsection{Task 5.1 Intent Classification}

\paragraph{Lynx (Rank 1)~\cite{lynx-nadi2026}.}
Team Lynx adopted a cascaded system utilizing \texttt{cohere-transcribe-arabic-07-2026} as a frontend to transcribe the utterances before applying extracting word and character-level n-gram features for TF-IDF representations. Character n-grams used n$\in \left\{2,5\right\}$, while they only used word unigrams and bigrams. These features are fused and provided to a one-rest logistic regression classifiers trained for each of the intents. To capture unknown intent scenarios, they also employ \texttt{MARBERTv2}~\cite{abdul2021arbert} model trained to for intent classification on the MASSIVE corpus~\cite{fitzgerald2023massive}, utterances that fall below a threshold probability for sum of the corresponding categories of the 5.1 subtask are labeled as unknown.

\paragraph{CIS-OpenSLU (Rank 2)~\cite{cis_openslu-nadi2026}.}
The system followed a two-stage pipeline in which \texttt{whisper-large-v3-turbo}, implemented with Faster-Whisper, was first used to transcribe the Tunisian Arabic speech. Intent recognition was then performed using a linear SVM trained on word- and character-level TF-IDF features extracted from the ASR transcripts. The official training and validation sets were combined, resulting in \(3{,}272\) labeled utterances, without external labeled data or synthetic augmentation. To handle intents not represented in the training data, confidence-based open-set recognition was applied using the maximum SVM decision score, with low-confidence predictions assigned to an \textit{unknown} class.

\paragraph{A3da2 Messi (Rank 3)
} 
The A3da2 Messi team did not submit a system paper.

\paragraph{Aslema (Rank 4)~\cite{aslema-nadi2026}.}
The system used \texttt{Qwen3-Omni-30B-A3B-Instruct} with joint LoRA fine-tuning for intent recognition and slot filling. Model selection was performed after evaluating several multimodal foundation models in both zero-shot and fine-tuned settings. The official training set, containing approximately \(2.7\)K examples, was augmented with synthetic Tunisian Arabic data generated using Gemini and filtered with an LLM-based quality check. Corresponding speech was synthesized using VoxCPM and further filtered for acoustic quality, yielding approximately \(23\)K additional speech--text pairs. The final model was trained on the combined original and synthetic data.

\paragraph{FARABI-Fraunhofer  (Rank 5)
} 
The FARABI-Fraunhofer  team did not submit a system paper.

\subsection{Task 5.2 Slot Filling}

\paragraph{Aslema (Rank 1)~\cite{aslema-nadi2026}.}
The system used \texttt{Qwen3-Omni-30B-A3B-Instruct} and jointly modeled intent recognition and slot filling through LoRA-based fine-tuning. Training combined the official dataset, containing approximately \(2.7\)K examples, with a substantially larger synthetic set. Synthetic Tunisian Arabic utterances with intent and slot annotations were generated using Gemini, filtered for quality, and converted to speech using VoxCPM~\cite{zhou2026voxcpm2}, resulting in approximately \(23\)K additional speech--text pairs. The final model was fine-tuned on the combined original and synthetic data, enabling direct prediction of slot labels and their corresponding values from speech.

\paragraph{Zila  (Rank 2)} 
The Zila team did not submit a system paper.


\end{document}